\documentclass[conference]{IEEEtran}
\usepackage{times}

\usepackage[numbers]{natbib}
\usepackage{multicol}
\usepackage[bookmarks=true]{hyperref}
\usepackage{xcolor}
\usepackage{amsmath}
\usepackage{graphicx, caption}
\usepackage{booktabs}

\newcommand{\method}{ArticuBot{}}

\begin{document}

    \title{\method: Learning Universal Articulated Object Manipulation Policy via Large Scale Simulation}




%
\author{\authorblockN{Yufei Wang$^{*1}$,
Ziyu Wang$^{*2}$,
Mino Nakura\authorrefmark{2}$^1$, 
Pratik Bhowal\authorrefmark{2}$^1$,
Chia-Liang Kuo\authorrefmark{2}$^3$, \\
Yi-Ting Chen$^3$,
Zackory Erickson\authorrefmark{3}$^1$,
David Held\authorrefmark{3}$^1$}
\authorblockA{$^1$Robotics Institute, Carnegie Mellon University}
\authorblockA{$^2$IIIS, Tsinghua University}
\authorblockA{$^3$Department of Computer Science, National Yang Ming Chiao Tung University }
\authorblockA{$^*$\authorrefmark{2}Equal Contribution, \authorrefmark{3}Equal Advising}
}

\makeatletter
\let\@oldmaketitle\@maketitle
\renewcommand{\@maketitle}{\@oldmaketitle
\centering
  \includegraphics[width=.96\linewidth]{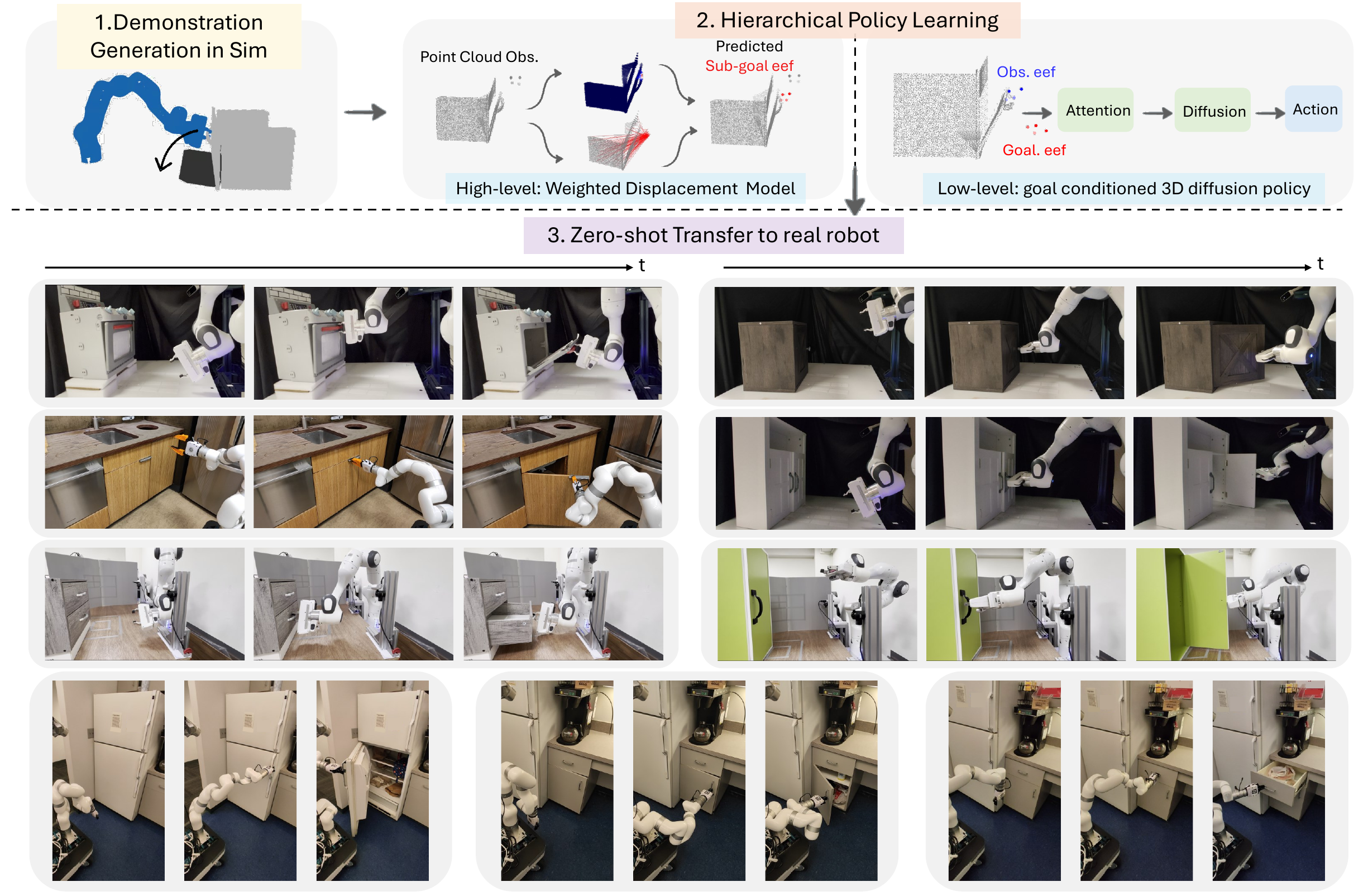}
  \small
  \vspace{-2.5mm}
  \captionof{figure}{
   Overview and real-world results of \method{}. 
   \textbf{Top}: We generate thousands of demonstrations using a physics-based simulator, and distill them into a hierarchical policy with point cloud observations. 
   \textbf{Bottom}: The learned policy can transfer zero-shot to table-top Franka arms in two different labs and a mobile X-Arm, and can open diverse unseen articulated objects in both labs, real kitchens and lounges.  
  }
  \label{fig:draw}
  \vspace{-3.5mm}
  }
\makeatother

\maketitle
\addtocounter{figure}{-1}


\begin{abstract}
This paper presents \method, in which a single learned policy enables a robotics system to open diverse categories of unseen articulated objects in the real world.  This task has long been challenging for robotics due to the large variations in the geometry, size, and articulation types of such objects. 
Our system, \method, consists of three parts: generating a large number of demonstrations in physics-based simulation, distilling all generated demonstrations into a point cloud-based neural policy via imitation learning, and performing zero-shot sim2real transfer to real robotics systems.
Utilizing sampling-based grasping and motion planning, our demonstration generalization pipeline is fast and effective, generating a total of 42.3k demonstrations over 322 training articulated objects.
For policy learning, we propose a novel hierarchical policy representation, in which the high-level policy learns the sub-goal for the end-effector, and the low-level policy learns how to move the end-effector conditioned on the predicted goal.  We demonstrate that this hierarchical approach achieves much better object-level generalization compared to the non-hierarchical version. 
We further propose a novel weighted displacement model for the high-level policy that grounds the prediction into the existing 3D structure of the scene, outperforming alternative policy representations. 
We show that our learned policy can zero-shot transfer to three different real robot settings: a fixed table-top Franka arm across two different labs, and an X-Arm on a mobile base, opening multiple unseen articulated objects across two labs, real lounges, and kitchens. 
Videos and code can be found on our project website: \href{https://articubot.github.io/}{https://articubot.github.io/}.

\end{abstract}

\IEEEpeerreviewmaketitle

\section{Introduction}

Robotic manipulation of articulated objects, such as cabinets, drawers, fridges, microwaves, has wide applications as such objects are ubiquitous in both industrial and household settings. Having a single robotics policy that can generalize to manipulate diverse articulated objects has long been challenging due to the large variations in the geometry, shape, size and articulation types of such objects. 
Many prior works have studied the problem of articulated object manipulation~\cite{xu2022universal, mo2021where2act, eisner2022flowbot3d, jain2021screwnet, jiang2022ditto, wu2021vat, gupta2024opening, morlans2024grasp}. However, few have demonstrated generalization to manipulating many different articulated objects in the real world without simplifying assumptions (e.g., using a suction gripper~\cite{eisner2022flowbot3d}). 
In this paper, we aim to learn a generalist articulated object manipulation policy that can open various kinds of articulated objects in the real world with commercial robotic manipulators equipped with a parallel jaw gripper, purely from visual observations, without assuming access to knowledge of the articulation parameters.  

Motivated by a recent trend of success in scaling up robot learning with large datasets, we aim to learn a universal articulated object opening policy following this paradigm: generating thousands of demonstrations in physics-based simulation, distilling the generated data into a generalizable policy by imitation learning, and then performing zero-shot sim2real transfer. 
This is a paradigm that has been applied in previous work to learn general policies for different robotics tasks, such as grasping~\cite{fang2023anygrasp, sundermeyer2021contact}, locomotion~\cite{lee2020learning, kumar2021rma, zhuang2023robot}, assembly~\cite{tang2024automate, tang2023industreal}, and deformable object manipulation~\cite{wang2023one, ha2022flingbot}. 

In this paper, we investigate various ways to realize such a system to learn a generalist policy for articulated object manipulation. 
We first build an efficient data generation pipeline that combines sampling-based grasping, motion planning, and action primitives. Using the pipeline, we have generated a large dataset consisting of thousands of (42.3k) demonstrations over 322 articulated objects. 
We also show that using a hierarchical policy representation, in which a high-level policy predicts sub-goal end-effector poses and a low-level policy predicts delta end-effector transformations, performs much better than the non-hierarchical version when imitating the generated large dataset.
We also explored various design choices for the policy representations to study which architecture scales up the best when learning with a large number of demonstrations. 
We show that a weighted displacement model that leverages the underlying 3d scene structure can scale and generalize  better  than models that do not incorporate such 3D reasoning. 

Our final policy, trained with 42.3k trajectories and 322 objects in simulation, can transfer zero-shot to the real world to open diverse unseen real articulated objects. 
Furthermore, although our policy is only trained on a Franka arm in simulation, we show that it can transfer zero-shot to two different embodiments in the real world: a table-top Franka arm, and a mobile-base X-Arm. 
This is achieved by using the policy to learn actions in the robotic arm's end-effector space instead of the joint space. 
Our final policy is successfully deployed in 3 different real-world settings: two table-top Franka arms in two different labs, and a mobile-base X-Arm in various real kitchens and lounges. 
This single policy is able to open 20 
different unseen real-world articulated objects such as cabinets, drawers, microwaves, ovens, and fridges in these different test settings, in a zero-shot manner. 
See Fig.~\ref{fig:draw} for a visualization of some of the different real-world articulated objects that our policy is able to open. 

In summary, our contributions are:
\begin{itemize}
    \item A system that presents a single policy trained on thousands of demonstrations generated in simulation, that can zero-shot transfer to the real world and generalize to open various articulated objects with 2 robot embodiments: a table-top Franka, and a mobile base X-Arm.  
    \item  We show that using a hierarchical policy representation is better than the non-hierarchical version to achieve object-level generalization. 
    \item We present a weighted displacement policy representation that scales up well with the number of demonstrations, outperforming alternative policy representations.
    \item A large articulated object manipulation simulation dataset that contains 42.3k demonstration trajectories for 322 articulated objects, and a pipeline for quickly generating additional demonstrations. 
\end{itemize}

\section{Related Work}

\subsection{Robot Learning for Articulated Object Manipulation}
There is a rich body of prior work studying the problem of articulated object manipulation~\cite{mo2021where2act, wu2021vat, eisner2022flowbot3d, zhang2023flowbot++, xu2022universal, morlans2024grasp, jain2021screwnet, zeng2021visual, pmlr-v164-jain22a, gupta2024opening, wang2022adaafford, xiong2024adaptive}. 
Most of these prior works show major results in simulation, with limited real-world manipulation results~\cite{mo2021where2act, wu2021vat, zhang2023flowbot++, xu2022universal, jain2021screwnet, zeng2021visual, pmlr-v164-jain22a, wang2022adaafford}. In contrast, our work aims to learn a manipulation policy that can transfer and generalize to diverse real-world articulated objects. 
\citet{eisner2022flowbot3d} shows a number of tests on real-world articulated objects in a table-top lab setting with a suction gripper to simplify grasping. Our policy works with the standard parallel jaw gripper which is more commonly equipped with robotic manipulators and perform grasping of the handles for opening. We also show results with a mobile manipulator in real kitchens and lounges. \citet{gupta2024opening} proposes a system that integrates various modules for perception, planning, and action and shows that it can open various cabinets and drawers with a mobile manipulator in real kitchens. Our method does not employ layered modules, instead, we directly learn a policy via imitation learning that maps sensory observations to actions. Our method also generalizes to a more diverse range of articulated objects such as fridges, microwaves, and ovens. 
\citet{xiong2024adaptive} builds a mobile base manipulator for articulated object manipulation and learn object-specific policies via imitation learning and reinforcement learning directly in the real world. We learn our manipulation policies by constructing much larger demonstration datasets in simulation and performing sim2real transfer, and we learn a single policy that can generalize to various articulated objects. 
Some  prior works learn to first predict the articulation parameters and then use the predicted articulation parameters for manipulation~\cite{jain2021screwnet, zeng2021visual, pmlr-v164-jain22a}. Our policy directly learns how to manipulate the object without explicitly inferring the articulation parameters. 
Another line of work~\cite{jiang2022ditto, chen2024urdformer, mandi2024real2code, liu2023paris} focus on reconstructing the articulated objects from real-world images to simulation. Our work focuses on manipulation rather than real2sim reconstruction. 
A recent work~\cite{morlans2024grasp} learns specific grasps for articulated objects that are useful for downstream manipulation. Our policy learns not only the grasping, but also the opening; further, we compare to this prior work and show significantly improved performance. 







\subsection{Sim2real Policy Learning}
Learning a policy via simulation training and then transferring to the real world (sim2real transfer) has been applied to many different domains in previous work, including legged locomotion~\cite{lee2020learning, cheng2024extreme, radosavovic2024real}, grasping~\cite{fang2023anygrasp, sundermeyer2021contact, dalal2024local}, in-hand object reorientation~\cite{akkaya2019solving, chen2023visual, qi2023general}, catching objects~\cite{zhang2024catch}, deformable object manipulation~\cite{wang2023one, xu2023roboninja}, and more~\cite{dalal2024neural, fishman2023motion}. No prior work has demonstrated the learning of a generalizable policy for articulated object manipulation via sim2real transfer. 
Many of these prior works use reinforcement learning and teacher-student learning to learn the policy in simulation~\cite{lee2020learning, radosavovic2024real, cheng2024extreme, qi2023general, zhang2024catch, wang2023one}. In contrast, we generate demonstrations in simulation using a combination of techniques including sampling-based grasping, motion planning, and action primitives, and learn the policy via imitation learning. 
Some recent work~\cite{wang2023robogen, wang2023gensim, hua2024gensim2} attempts to automate simulation policy learning for many tasks. In contrast, our work focuses on learning a generalizable policy for articulated object manipulation. 



\subsection{Robotic Foundation Models}
Many recent works aim to develop a foundation model for robotics, where a single model can perform multiple tasks or generalize to different settings~\cite{brohan2022rt, brohan2023rt, kim2024openvla, team2024octo, etukuru2024robot, o2023open, dalal2024local, lin2024data}. 
Most of them perform imitation learning with a large set of demonstrations collected in the real world~\cite{brohan2022rt, brohan2023rt, kim2024openvla, team2024octo, etukuru2024robot, o2023open, lin2024data}. Instead, we generate demonstrations and learn the policy in simulation, and then we perform sim2real transfer to deploy it in the real world. Most of these works do not focus on tasks involving articulated objects and do not demonstrate the policy working for manipulating diverse articulated objects~\cite{brohan2022rt, brohan2023rt, kim2024openvla, team2024octo, lin2024data, dalal2024local}, while our paper focuses on building a generalizable policy specifically for articulated object manipulation. \citet{etukuru2024robot} shows the most diverse real-world test settings for articulated object manipulation among these previous works. 
However, they train two different policies for drawer and cabinet opening. In contrast, we train a single model that can be applied to opening various categories of articulated objects. Besides, their system requires a specialized gripper, whereas our method works for general parallel jaw grippers and transfer across two different grippers. 


\section{Problem Statement and Assumptions}
The task we are considering is for a robotic arm to open an articulated object within the category of drawers, cabinets, ovens, microwaves, dishwashers, and fridges. We assume that the object should have a graspable handle, so it can be opened in the fully closed state. We aim to learn a policy $\pi$, that takes as input a sensory observation and robot proprioception $o$, and outputs actions $a$ that opens the articulated object. 
We assume the robot arm is equipped with a common parallel jaw gripper instead of a suction gripper or a floating gripper, which are often assumed in prior works for simplification~\cite{eisner2022flowbot3d, xu2022universal}.  
We also assume access to a pool of articulated object assets to be used in simulation, as well as annotations to handles (in simulation only). 
For effective sim2real transfer, we use point clouds as the sensory observations.
We assume the name of the target object to manipulate, such that we can run a open-vocabulary segmentation method, e.g., Grounded SAM~\cite{ren2024grounded}, to segment the object and obtain object-only point clouds.

\begin{figure*}
    \includegraphics[width=\linewidth]{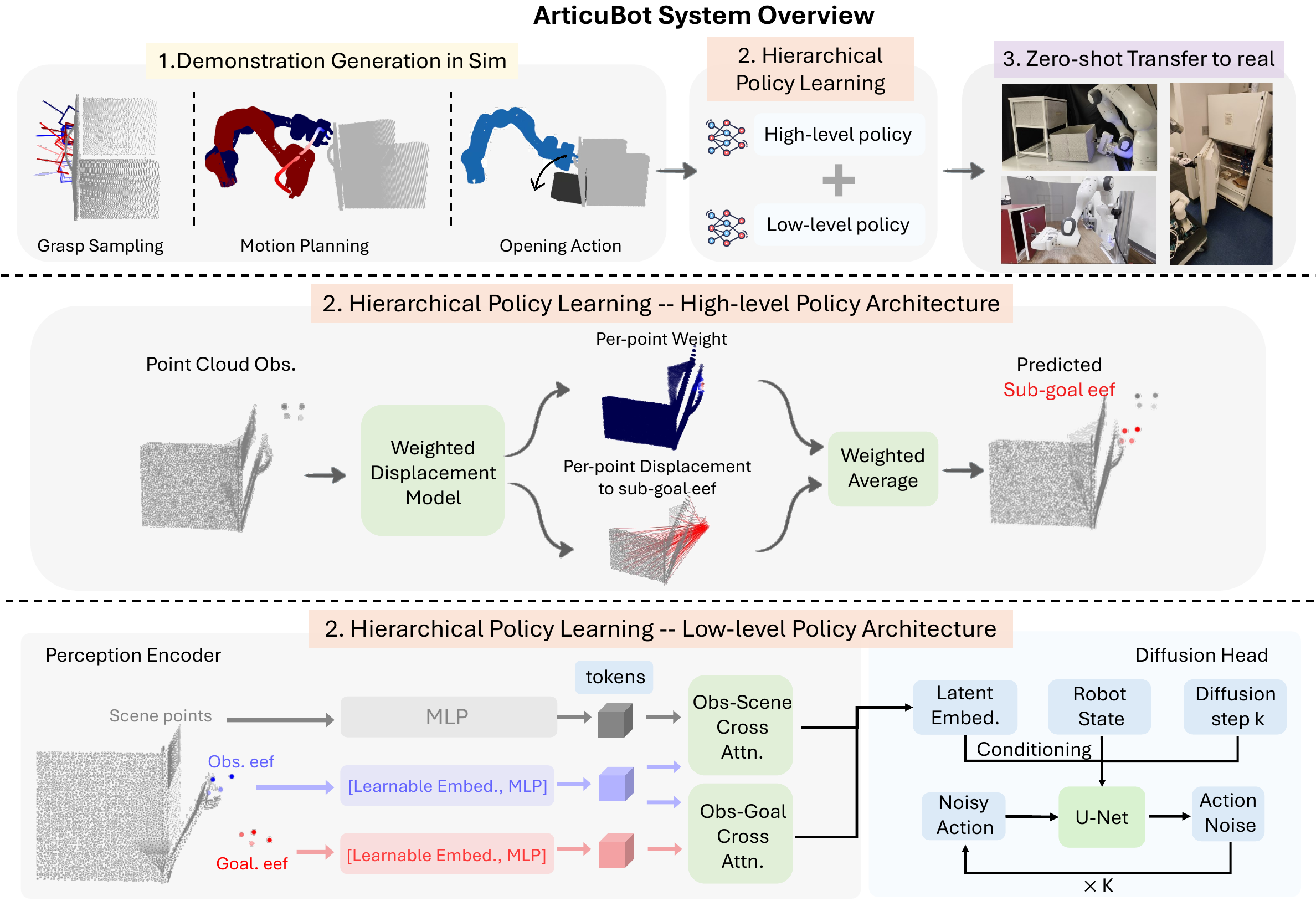}
    \vspace{-5mm}
    \caption{System overview of \method{}. 
    \textbf{Top}: We combine sampling-based grasping, motion planning, and opening actions to efficiently generate thousands of demonstrations in simulation. These demonstrations are distilled into a hierarchical policy via imitation learning, and then zero-shot transferred to real world.
    \textbf{Middle}: We propose a weighted displacement model for the high-level policy, which predicts the sub-goal end-effector pose. The weighted displacement model predicts the displacement from each point in the point cloud observation to the sub-goal end-effector, as well as a weight for each point. The final prediction is the weighted average of each point's prediction.
    \textbf{Bottom}: We propose a goal-conditioned 3D diffusion policy for the low-level policy, which first applies attention between the current end-effector points, the scene points, and the goal end-effector points to obtain a latent embedding, and then performs diffusion on the latent embedding to generate the action, which is the delta transformation of the robot end-effector.}
    \vspace{-5mm}
    \label{fig:system}
\end{figure*}

\section{\method{}}
Fig.~\ref{fig:system} gives an overview of our system, which consists of 3 stages. The first is large-scale demonstration generation, in which we combine methods from motion planning, sampling-based grasping, and action primitives to generate thousands of demonstrations in simulation. The second is hierarchical policy learning, in which we perform imitation learning on the generated demonstrations to distill them into a vision-based policy. Finally, we deploy our simulation-trained policy zero-shot to the real world, on two table-top Franka arms in two different labs and an X-Arm on a mobile base in real kitchens and lounges, opening real-world cabinets, drawers, microwaves, fridges, dishwashers, and ovens. 

\subsection{Demonstration Generation in Simulation}
\label{sec:demon_generation}
First we describe our procedure for automatically generating thousands of demonstrations in simulation.
We use the PartNet-Mobility~\cite{xiang2020sapien} dataset, which contains hundreds of articulated objects. Among these, we use the categories of storage furniture, microwave, oven, dishwasher, and fridge. The majority of the assets in these categories have annotations of handles; we filter out assets that do not have such annotations, since the position of the handle is needed for generating the demonstrations. 

The process of opening an articulated object can be decomposed into two substeps: grasping the handle, and then opening it along the articulation axis. 
Our demonstration generation pipeline follows these two substeps as well (see Fig.~\ref{fig:system} top left for an illustration of the process):
we first perform sampling-based grasping to generate hundreds of end-effector grasping poses on the handle. 
For each generated grasp, we approach the grasping pose using collision-free motion planning. 
After performing the grasp, since we have the ground-truth articulation information of the object in simulation, we move the end-effector along the articulation axis for a fixed distance to open it. 
We detail each of these three steps below. 

\noindent\textbf{Simulation Initialization: } We use a Franka arm in simulation for generating the demonstrations. The base of the Franka Arm is initialized at the world origin. We randomize the position, orientation, and size of the object, as well as the initial joint angle of the Franka Arm, to increase diversity in the generated demonstrations. The detailed parameters for the randomization can be found in Appendix~\ref{app:simulation-randomization-details}. 

\noindent\textbf{Sampling Based Grasping: }Given an articulated object from PartNet-Mobility and a link (i.e., a door) we want to open, we first obtain a point cloud of the link's handle using the annotations from the dataset. We perform farthest point sampling on the handle point cloud to get $m_1=15$ candidate grasping positions. For each grasping position, to generate the grasping orientation, we align the z-axis of the robot end-effector (which is the direction that points from the root of the hand to the finger) with the normal direction of that handle point. 
We set the y direction of the end-effector (which is the direction along which the finger opens and closes) to be horizontal if the handle is vertical (i.e., its height is larger than its width), and vice versa. We also sample $m_2=8$ random small angle perturbations ($<30^\circ$) about the y axis  to increase the diversity  of our grasp pose candidates. This generates in total $m_1\times m_2=120$ grasping pose candidates. See Fig.~\ref{fig:system} (top left) for an illustration of the sampled grasps. 

\noindent\textbf{Motion Planning for reaching the grasping pose:} For each of the grasp candidates, we first use inverse kinematics (IK) to compute a target joint configuration of the robot arm. We solve the IK for $m_3=80$ times and filter out solutions that have collisions between the robot arm and the environment (e.g. collisions with the floor or the target object). Among the collision-free solutions, we choose the one solution that has the shortest distance in the joint angle space to the current joint configuration, so as to minimize the distance of the path needed to reach the target joint configuration.  
We then run three different motion planning algorithms, RRT*~\cite{karaman2011sampling}, BIT*~\cite{gammell2015batch} and ABIT*~\cite{strub2020advanced}, to generate the path to reach the target configuration. 
We smooth the resulting path from each algorithm by shortcutting unnecessary waypoints and using B-spline smoothing~\cite{hauser2010fast}.
We keep the path that has the shortest length in terms of the total end-effector movement. See Fig.~\ref{fig:system} (top left) for a visualization of the motion planned path. 

\noindent\textbf{Generating Opening Actions:} 
Next, we generate demonstrations in simulation of the robot executing the opening action. After the grasping pose is reached via motion planning, we close the gripper to form a grasp.
Using the ground-truth articulation information of the object, we can compute an idealized end-effector trajectory that opens the object perfectly. Formally, let $T^{init}_{eef}$ represent the end-effector's pose after it grasps the handle, and let $T_{door}(\theta)$ represent the pose of the door at joint angle $\theta$. 
We compute the idealized trajectory based on the fact that the relative pose between the robot end-effector and the door should remain unchanged during the trajectory of opening the door, i.e., $T_{rel} = T^{-1}_{door}(\theta) T_{eef}$ should be a constant for any joint angle $\theta$. 
Assume the door is at joint angle $\theta_{init}$ when the robot grasps it, then the pose of the robot end-effector when the door is opened at joint angle $\theta$ can be computed as: $T_{eef} = T_{door}(\theta) T^{-1}_{door}(\theta_{init}) T^{init}_{eef}$. 
We can then compute a trajectory for the end-effector pose that opens the object with increasing values of $\theta$, e.g., from $0^\circ$ to $90^{\circ}$ with an interval of $1^{\circ}$.
IK is then performed for the end-effector to reach each of the computed poses along the trajectory to open the object. 


\noindent\textbf{Filtering: } Some of the trajectories will fail to fully open the door due to various reasons such as: no collision-free joint angles can be found at the sampled grasping pose, motion planning failed to find a collision-free path to reach the grasping pose, the grasping pose does not result in a firm grasp of the handle, or the end-effector slips off the handle partway during opening. 
We filter out all trajectories where the final opened angle (radians for hinge doors and centimeters for drawers) is smaller than a threshold, e.g., if the door is opened less than 60 degrees. 
From the remaining successful trajectories, we choose the single best trajectory according to the following two metrics: 1) the stability of the grasp, which is approximately measured as the number of handle points that are between the end-effector fingertips, and 2) the length of the motion planned trajectory (in the end-effector space) to reach the grasping pose. 
Each trajectory is first ranked using these two metrics, and the final rank is the sum of the two individual ranks. The trajectory with the highest rank is kept as the final best trajectory for opening the door.

By employing the above data generation pipeline, and executing each of the $m_1 \times m_2$ trials in parallel, we can generate optimal trajectories for opening an articulated object. Using a CPU with 128 virtual cores, one optimal opening trajectory can be generated within 2 minutes. 
Using this approach, we have generated 42.3k successful opening trajectories for 322 objects in PartNet-Mobility.

\subsection{Policy Learning with a Hierarchical Policy Representation}
\label{sec:method-policy}
We now describe how we distill the above generated trajectories into a vision-based neural policy via imitation learning. 
Formally, the above approach gives us a set of generated demonstrations $\{\tau_i\}_{i=1}^N$, where each trajectory $\tau_i$ is a list of observation-action pairs: $\tau_i = \{(o_1^i, a_1^i), ..., (o_T^i, a_T^i)\}$. The observations include point clouds of the scene and the robot proprioception (end-effector pose and finger open/close). 
We perform segmentation on the scene point cloud to remove the background and leave only the target object.  
In simulation, this can be achieved using the ground-truth segmentation masks provided by the simulator; in the real world, we use an open-vocabulary object segmentation model, e.g., Grounded SAM~\cite{ren2024grounded}. 
See Fig.~\ref{fig:system} (middle) for an example object point cloud in simulation. 
The actions represent the delta transformation of the end-effector, which includes the delta translations, delta orientations, and delta finger movement. 
We use the robot base frame as our reference frame, i.e., all point cloud observations, and robot actions, are expressed in the robot base frame. 

Our goal is to find a neural network policy $\pi$, parameterized by $\theta$, to minimize the following imitation learning loss:
\begin{equation}
\small
    \mathcal{L}_{\theta} = \sum_{i=1}^N \sum_{t=1}^T || \pi_\theta(o^i_t) - a^i_t ||_2^2
\end{equation}

The goal for the policy is to be able to generalize to open various kinds of different objects, which possess diverse geometries, shapes, and articulation types.  We hypothesize that, to achieve object-level generalization in such a case, it is inefficient to just learn to predict actions as the low-level delta transformations of the end-effector. Instead,  we propose to use a hierarchical policy representation, which consists of a high-level policy and a low-level policy. The high-level policy will learn to predict the sub-goal end-effector poses, e.g. intermediate waypoints of where the gripper should be at various key frames in the trajectory. The low-level policy still learns to predict the low-level delta transformations of the end-effector at each timestep, but it is additionally conditioned on the high-level prediction of the sub-goal end-effector pose, which helps the low-level policy to better generalize across diverse objects. 
We now detail how each of the policies work.   

\noindent\textbf{High-Level Policy. }
Intuitively, the high-level policy aims to predict ``where'' the robot should move to. 
Specifically, the high-level policy $\pi_\theta^H$ learns to predict the sub-goal end-effector pose given an observation. The sub-goal end-effector pose is defined as the pose of the robot end-effector at the end of each substep for a given task. In our case, the task of opening an articulated object (e.g., a cabinet) can be decomposed into two substeps: grasping the handle and opening the door. 
Thus for this task, the sub-goal end-effector poses are  the poses where the robot has grasped the handle, and when it has fully opened the door. 
Formally, the high-level policy $\pi^H_\theta$ is learned via minimizing the following loss:
\begin{equation}
\small
     \mathcal{L}_{\theta} = \sum_{i=1}^N \sum_{t=1}^T || \pi^H_\theta(o^i_t) - a^i_{pose_t} ||_2^2
\end{equation}
where $a^i_{pose_t}$ is the sub-goal end-effector pose at timestep $t$, which is represented as its 3D position, orientation, and the gripper finger opened width. 

We propose a new representation for the high-level policy, termed the \textit{weighted displacement model}. 
Existing 3D neural policies, e.g., Perceiver-Actor~\cite{shridhar2023perceiver}, or 3D Diffuser Actor~\cite{ke20243d}, often generate the sub-goal end-effector pose in free SE(3) space.
Instead, we aim to predict the sub-goal end-effector pose by grounding the prediction on the observed 3D structure of the scene. To do so, we design the policy to learn to predict the ``offset'' from observed points in the scene to the sub-goal end-effector pose. This learned offset thus closely grounds the prediction in the observed 3D scene structure.
See  Fig.~\ref{fig:system} (middle) for an overview of the weighted displacement model.

Specifically, instead of representing the sub-goal end-effector pose as a position and an $SO(3)$ orientation (e.g., a quaternion or a 6D orientation representation~\cite{zhou2019continuity}) and forcing the network to learn the connection between $SO(3)$ orientations and the 3D point cloud observation, we propose to represent the sub-goal end-effector pose as a collection of $K$ points that are naturally in 3D.
In our case, we use $K=4$: the first point is located at the root of the robot hand, the second and third points at the parallel jaw gripper fingers, and the fourth point at the grasping center when the finger closes. 

In this way, a sub-goal end-effector pose can be represented as $\{ee_i\}_{i=1}^4$, where $ee_i$ is the 3D position of the $i^{th}$ point. 
Given a point cloud of the scene with $M$ points $P = \{p_j\}_{j=1}^M$ and the current robot end-effector points $\{ee^{obs}_i\}_{i=1}^4$, we propose to let the policy $\pi^H_\theta$ learn to predict the displacement from each point $p_j$ in the scene point cloud to the sub-goal end-effector points $\{ee^{goal}_i\}_{i=1}^4$: $\delta_j = [\delta_j^1, \delta_j^2, \delta_j^3, \delta_j^4]$, where $\delta_j^i = ee^{goal}_i - p_j$. 
At inference time, the final predicted sub-goal end-effector pose is the averaged prediction from all points in the scene: $ee_{i}(\theta) = \sum_{j=1}^M (p_j + \delta_j^i(\theta))$. 
This proposed model converts the prediction of the end-effector pose from $SE(3)$, especially $SO(3)$, to a list of vectors just in the 3D space, which is less complex.
We note this per-point prediction requires us to use a network architecture that can generate per-point outputs given a point cloud input. Many point cloud processing networks can do so~\cite{qi2017pointnet, qi2017pointnet++, zhao2021point}; we choose to use PointNet++~\cite{qi2017pointnet++} in our case.  
As the model predicts the displacement from existing points in the scene instead of the absolute positions, and PointNet++ is a translation-invariant architecture, our proposed model is thus invariant to the translation of the robot end-effector and the object, which makes it more robust in real-world settings. 


However, not all points in the scene are of equal importance for the task and for predicting the sub-goal end-effector pose. 
In the task of opening an articulated object (e.g., a cabinet), the points on the handle are probably more important compared to the points on the side of the cabinet. 
Therefore, we propose for the network to also learn a weight for each point in the scene point cloud when predicting the sub-goal end-effector pose. 
Formally, in addition to predicting the per point displacement $\delta_j$, the network also predicts a weight $w_j$ for each point $p_j$. 
At inference time, the final prediction of the sub-goal end-effector points is then the weighted average of the displacement from each point: 
$
ee_i(\theta) = \sum_{j=1}^M w_j(\theta) (p_j + \delta_j^i(\theta) ), i=1,2,3,4.
$

We term this high-level policy representation the \textit{weighted displacement model}. We train it with the following two losses, which supervises the per-point displacement prediction, and the weighted average prediction: 

\begin{align}
\vspace{-2mm}
\small
\label{eq:high-level-loss}
    L = \lambda_1 \frac{1}{M} \sum_{j=1}^M ||\delta_j &- \delta_j(\theta)||_2^2 + \lambda_2 \frac{1}{4} \sum_{i=1}^4 || ee^{goal}_i - ee_i(\theta) ||_2^2 \\
    ee_i(\theta) &= \sum_{j=1}^M w_j(\theta) (p_j + \delta_j^i(\theta) )
\end{align}

\noindent\textbf{Low-level Policy.}
The low-level policy $\pi^L_\theta$ learns to predict the delta transformation of the end-effector, given the observation $o$ and the sub-goal end-effector pose $\{ee^{goal}_i\}_{i=1}^4$, i.e., ``how'' to actually move the end-effector to solve the task. It is learned to minimize the following loss:
\begin{equation}
\small
    \mathcal{L}_{\theta} = \sum_{i=1}^N \sum_{t=1}^T || \pi^L_\theta(o^i_t, \{ee^{goal}_k\}_{k=1}^4\}) - a^i_t ||_2^2,
\end{equation}
where $a^i_t$ is the delta transformation of the end-effector, including the delta translation, delta rotation, and delta finger movement (open/close). 
We represent the delta rotation using the 6D rotation representation~\cite{zhou2019continuity}. 
We note that the low-level policy is not trained to reach the sub-goal end-effector pose; it is trained to solve the task, and the sub-goal end-effector pose is just an additional input that helps guide the low-level policy to learn how to move. 
Given that part of the demonstration trajectories are generated from a motion planner, which can be highly multi-modal, we employ a diffusion policy as the low-level policy representation, which is known for their ability to handle multi-modalities. 

Specifically, we modify 3D Diffusion Policy (DP3)~\cite{ze20243d} such that it can be conditioned on the sub-goal end-effector pose. 
See Fig.~\ref{fig:system} (bottom) for an overview of the low-level policy architecture. 
As in DP3, the network has two parts: a point cloud encoder that encodes the point cloud observation into a latent embedding, and a diffusion head on the latent embedding that generates the actions. 
We modify the encoder architecture to incorporate the sub-goal end-effector pose. 

Formally, given the current scene point cloud observation $P=\{p_j\}_{j=1}^M$, the current end-effector pose represented with 4 points $\{ee^{obs}_i\}_{i=1}^4$, the sub-goal end-effector pose $\{ee^{goal}_i\}_{i=1}^4$, we treat each point as a token and perform attention among them to generate the final latent embedding. 
For the scene point cloud $P=\{p_j\}_{j=1}^M$, we use an MLP applied to each point in the point cloud to obtain a per-point feature $\{f_j\}_{j=1}^M$, which will be used as the features for cross attention later. 
For the current end-effector points $\{ee^{obs}_i\}_{i=1}^4$, its feature for attention includes the following: the first part is a learnable embedding $v^{obs}_i$ for each of the 4 points. The second part is a feature vector produced by an MLP, where the input to the MLP includes each point's position $ee^{obs}_i$, the displacement to the corresponding point in the sub-goal end-effector pose $\delta_i = ee^{goal}_i - ee^{obs}_i$, and the displacement to the closest scene point: $\delta'_i = p_k - ee^{obs}_i, k = \arg\min_j ||p_j - ee^{obs}_i||$. The final feature vector for each point is  $f^{obs}_i = [v_i, \text{MLP}^{obs}(ee^{obs}_i, \delta_i, \delta'_i)]$. 
Intuitively, the displacement to the corresponding sub-goal points help the model to learn how to reach towards the goal; and the displacement to the closest scene points help the model to learn to avoid collision. 

We then perform cross attention between the scene point cloud and the current end-effector points with Rotary Position Embedding (RoPE)~\cite{su2024roformer}, which generates the updated features for current end-effector points as $\{f^{\text{obs-scene}}_i\}_{i=1}^4$. 
We generate the features for the goal end-effector points in the same way as for the current end-effector points: $f^{goal}_i = [v^{goal}_i, \text{MLP}^{goal}(ee^{goal}_i, \delta_i, \delta'_i)]$. 
We perform cross attention between the current end-effector points and the goal end-effector points, also with Rotary Position Embedding (RoPE)~\cite{su2024roformer}, which produces another set of updated features for the current end-effector points $\{f^{\text{obs-goal}}_i\}_{i=1}^4$. 
The final latent embedding used for diffusion is the concatenation of the above two features: $[f^{\text{obs-scene}}_{1}, f^{\text{obs-goal}}_{1}, ..., f^{\text{obs-scene}}_{4}, f^{\text{obs-goal}}_{4}]$. 
This latent embedding is used as the conditioning for an action generation UNet diffusion head, which takes as input this latent conditioning, the robot state (which includes the 3D position, 6D orientation of the end-effector, and finger width), a noisy version of the action, a denoising time step,  and predicts the noise. At test time, we use DDIM~\cite{song2020denoising} as the denoising scheduler to generate the actions. 


\begin{table}[t]
\scriptsize
\centering
\begin{tabular}{c|ccccc}
\toprule
& 10 objs & 50 objs & 100 objs & 200 objs & 322 objs 
 \\ \midrule
\begin{tabular}[c]{@{}c@{}}With  \\ camera randomization  \end{tabular} & 
\begin{tabular}[c]{@{}c@{}}1116  \\ 121k  \end{tabular}
& \begin{tabular}[c]{@{}c@{}}4656  \\ 502k  \end{tabular} & 
\begin{tabular}[c]{@{}c@{}} 8749   \\ 958k  \end{tabular}&  
\begin{tabular}[c]{@{}c@{}} 17893   \\ 1.97M  \end{tabular}&
\begin{tabular}[c]{@{}c@{}} 22918   \\ 2.55M  \end{tabular}\\  \midrule

\begin{tabular}[c]{@{}c@{}}Without  \\ camera randomization  \end{tabular} & 
\begin{tabular}[c]{@{}c@{}}750   \\ 80k  \end{tabular} 
& \begin{tabular}[c]{@{}c@{}}3669   \\ 403k  \end{tabular} & 
\begin{tabular}[c]{@{}c@{}} 6444   \\ 702k  \end{tabular}&  
\begin{tabular}[c]{@{}c@{}} 11795   \\ 1.28M  \end{tabular}&
\begin{tabular}[c]{@{}c@{}} 15998   \\ 1.76M  \end{tabular} \\
\bottomrule
\end{tabular}
\caption{Dataset Statistics. Top: \# of trajectories. Bottom: total \# of observation-action pairs in the trajectories. }
    \vspace{-5mm}
\label{tab:dataset_statistics}
\end{table}

\subsection{Zero-shot Transfer to Real Robotic Systems}
\label{sec:zero-shot-transfer}
After the high-level and low-level policies are trained in simulation, we transfer them zero-shot to real-world robotic systems. 
During inference, at each time step, given the current point cloud observation and end-effector pose, we first run the high-level policy to obtain a predicted goal end-effector pose, and then run the low-level policy with the point cloud observation, current end-effector pose, and the predicted goal end-effector pose to move the end-effector. We repeat this process until the object is fully opened, or a pre-defined episode length is reached, or the robot is going to collide with the environment.  
We discuss the details of our robot systems and real-world pipeline in Sec.~\ref{sec:real-world-experiments}.


\section{Simulation Results}
\subsection{Experiment Setups}
We use Pybullet~\cite{coumans2021} as the underlying physics simulator; any simulator that supports rigid-body dynamics and fast parallelization can be used. We use the PartNet-Mobility~\cite{xiang2020sapien} dataset for the assets of the articulated object. 
We extracted 332 objects from 5 different categories: storage furniture, microwave, dishwashers, oven, and fridge, which have annotations for handles. Among these, 322 are used for training and 10 unseen objects are used for testing. 
For each object, we generate 75 demonstrations for opening it. Each demonstration has a different configuration, where we vary the position, orientation, size of the object, and the initial pose of the end-effector (randomization details in Appendix~\ref{app:simulation-randomization-details}).


In order to study the object-level generalization abilities of different methods, we first generated 15,998 demonstration trajectories with 1.76M observation-action pairs for these 322 training objects, without any camera randomizations when rendering the point clouds. 
For efficient sim2real transfer, we generate additional demonstrations with camera pose randomizations. The datasets with camera randomizations has in total 22,918 trajectories and 2.55M observation-action pairs. 
We partition both types of datasets into different sets, in which we vary the number of objects in each of these sets (objects and trajectories are randomly sampled) to study the scaling behavior of different methods. 
The detailed statistics of the partitioned datasets can be found in Table~\ref{tab:dataset_statistics}.

For evaluation, we test each of the 10 objects with 25 different configurations, resulting in a total of 250 test scenarios. 
The evaluation metric is the \textbf{normalized opening performance}: the ratio of the increase in the opened joint angle of the object achieved by a method, to the increase in the opened joint angle of the object in the demonstration, which is calculated as \( \frac{\theta_{f} - \theta_{0}}{\theta_{demo} - \theta_{0}} \), where \( \theta_{0} \) is the initial opened angle, \( \theta_{f} \) is the final opened angle achieved by the method and \( \theta_{demo} \) is the final opened angle in the demonstration. A value of 1 indicates that the method performs as well as the demonstration, while 0 means the method has not contributed to opening the object.
For each method, we run the evaluation 3 times (a total of 750 trials) and report the mean and standard deviation of the normalized opening performances of the 3 runs.  
In the following, we compare to different baselines and prior methods to answer different research questions. 


\begin{figure}[t]
    \centering
    \includegraphics[width=.95\linewidth]{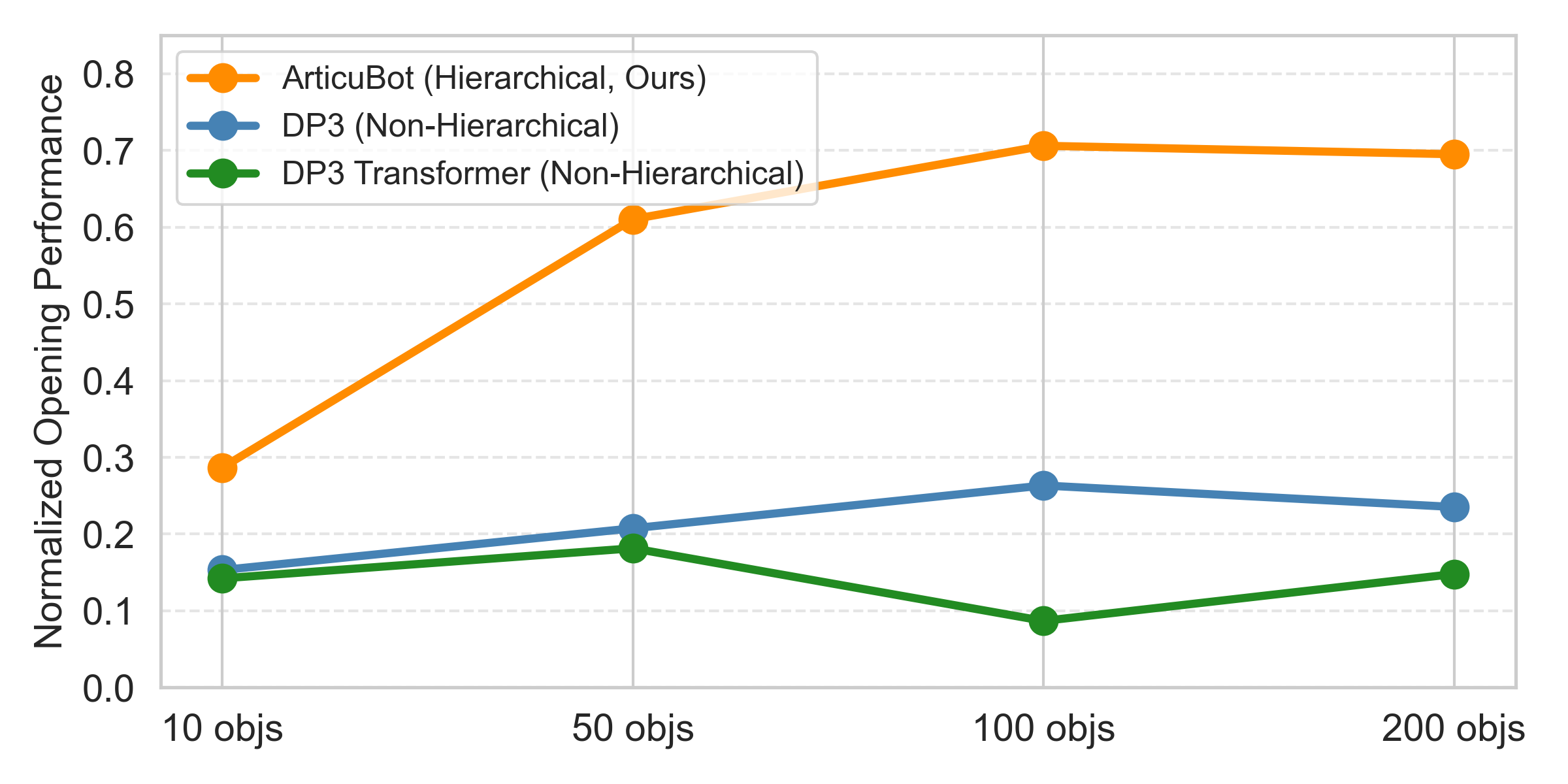}
    \vspace{-3mm}
    \caption{Comparison of hierarchical and non-hierarchical policies. }
    \vspace{-5mm}
    \label{fig:hierarchical-line-plot}
\end{figure}

\begin{figure*}[t]
    \centering
    \includegraphics[width=\linewidth]{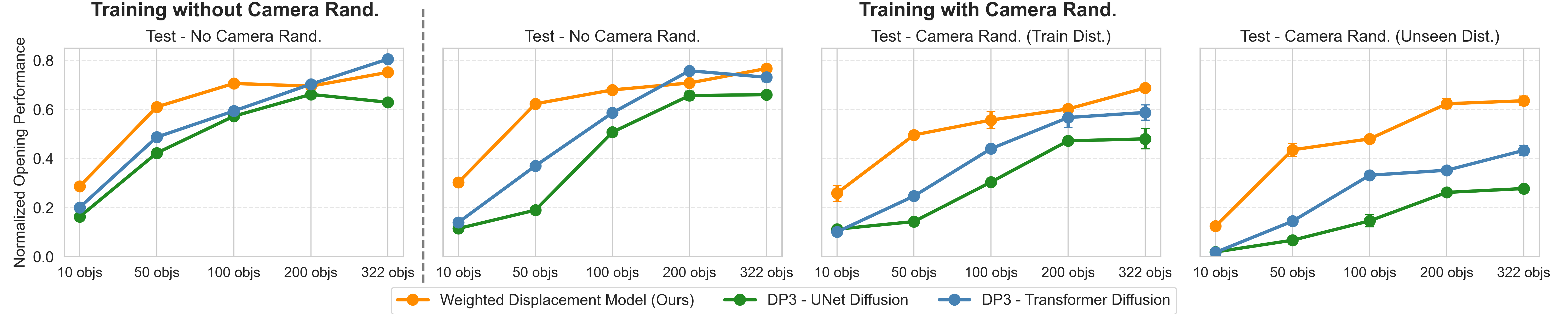}
    \vspace{-5mm}
    \caption{Comparison of different high-level policies. Leftmost: Train and test without camera randomizations. Right: Train with camera randomizations, and test with no camera randomization, with camera randomizations from training distribution, and with camera randomizations from an unseen test distribution.}
    \label{fig:high-level-cam-line-plot}
    \vspace{-5mm}
\end{figure*}


\subsection{Is a Hierarchical Policy Needed?}
Our first set of experiments aims to answer whether it is beneficial to use a hierarchical policy. 
We compare our proposed \textbf{hierarchical policy} with the following two non-hierarchical baselines: 
\begin{itemize}
 \item \textbf{3D Diffusion Policy (DP3)}~\cite{ze20243d}, a diffusion policy that takes 3D point cloud as input and outputs delta end-effector transformations as the actions. 
\item \textbf{DP3 Transformer}, which replaces the simplified PointNet encoder in DP3 with a transformer-based encoder (the same one used in our low-level policy in Sec.~\ref{sec:method-policy}). 
\end{itemize}
We compare these two baselines with our method that uses a hierarchical policy on datasets without camera randomizations to study the object-level generalization abilities of them.  
The results are shown in Fig.~\ref{fig:hierarchical-line-plot}. 
As shown, the performance of using a non-hierarchical policy only gets a normalized opening performance below 0.25, which is much lower than that of using a hierarchical policy.
Furthermore, we observe that the non-hierarchical policies do not experience significant improvement in performance as the number of training objects and trajectories increase. 
These results show that it is very challenging to achieve object-level generalization if we just learn low-level delta end-effector transformations, regardless of how much training data we use; using a hierarchal policy achieves much better object-level generalization performances. 

\subsection{Comparison of Different High-level policies}

We now investigate the performance of different high-level policy architectures. We compare our proposed \textbf{Weighted Displacement Model} to the following baselines:
\begin{itemize}
   \item \textbf{DP3 - UNet Diffusion}: this baseline builds upon DP3 and diffuses the sub-goal end-effector points. We modify the simplified PointNet encoder in DP3 to an attention-based encoder (similar to our low-level policy), as we find this provides better performance in early experiments.
    \item \textbf{DP3 - Transformer Diffusion}: In addition to using the attention-based encoder, we also modify the UNet diffusion head in DP3 to be a transformer-based architecture, such that the diffusion head conditions on not only a latent embedding, but also the 3D point cloud features. 
    \item  \textbf{3D Diffuser Actor (3DDA)}~\cite{ke20243d}: this baseline also diffuses the sub-goal end-effector points conditioned on 3D point cloud features, but employs a different architecture compared to DP3 - Transformer Diffusion.
\end{itemize}
Please see Appendix~\ref{app:high-level-detail} for more details about these baselines. 

We use a fixed low-level policy for all experiments in this section.
We first compare all method's performances when trained on the datasets without camera randomizations. 
The results are shown in the left subplot of Fig.~\ref{fig:high-level-cam-line-plot}. As shown, our proposed weighted displacement model performs consistently better than other methods when the number of training objects ranges from 10 to 100. When training with all 322 objects, DP3 - Transformer Diffusion achieves the best performance, outperforming the weighted displacement policy by 5\%. 
We also find that all methods' performances generally improve as the size of the dataset increases (except for DP3 - UNet diffusion when the number of training objects increases from 200 to 322, and weighted displacement model when the number of training objects increases from 100 to 200). 
We trained 3DDA with 200 objects and found it to perform poorly, achieving a performance of only 0.135, much lower than the performance of alternative methods ($> 0.6$). Therefore, we omit the training of it on other datasets to save computation.

Since our primary focus is sim2real transfer of the learned policy to the real world, we also compare these methods on the dataset with camera randomizations, as it is hard to place the camera at the exact pose in the real world as in simulation, and we want the policy to be robust to camera pose changes. 
For evaluation, we have three different settings: test on a fixed camera pose, test on random camera poses sampled from the training distribution, and test on random camera poses sampled from a test distribution not seen during training. 
The results are shown in the right 3 subplots in Fig.~\ref{fig:high-level-cam-line-plot}. 
Interestingly,
we find that when tested with camera randomizations, our proposed weighted displacement model performs much better than the compared methods, for all different sizes of training datasets. 
The performance gap is especially large when tested with unseen camera randomizations. 
Although DP3 - Transformer still achieves good performance when tested with a fixed camera pose with datasets more than 200 objects, its performance degrades drastically when tested with randomized cameras. In contrast, the performance drop for weighted displacement model when tested under camera randomizations is much smaller. 
We also find DP3 - UNet diffusion to perform poorly in this setting, which could be due to that it compresses the 3D scene into a single latent embedding vector, losing some of the needed 3D information for making the prediction. 
Similarly, we find that training with more data is generally helpful for achieving a higher performance.

\begin{table}[t]
\scriptsize
\centering
\begin{tabular}{c|c}
\toprule
 Ablations (trained with 200 objs) & \begin{tabular}[c]{@{}c@{}}Normalized \\Opening Performance\end{tabular} \\ \midrule
\method~(Ours) & $\mathbf{0.7 \pm 0.01}$ \\
Weighted Displacement Model w/ Point Transformer & $0.6 \pm 0.02$ \\
Unweighted Displacement Model & $0.66 \pm 0.01$ \\
Weighted Displacement Model w/ 6D orientation & $0.53 \pm 0.01$ \\ 
Replacing low-level policy with a motion planner & $0.24 \pm 0.02$ \\
\bottomrule
\end{tabular}



\caption{Performance of different ablation studies.}
\vspace{-6mm}
\label{tab:ablation}
\end{table}

\subsection{Ablation Studies}
\label{sec:ablation}
In this subsection, we examine some of the design choices in our method to understand their contributions. We compare our full method to the following ablations: 
\begin{itemize}
    \item \textbf{Weighted Displacement Model w/ Point Transformer~\cite{zhao2021point}}: instead of using PointNet++, we use the Point Transformer architecture for the weighted displacement high-level policy. 
    \item \textbf{Unweighted Displacement Model}: This ablation does not learn a weight for each point in the weighted displacement model; instead, the prediction is simply the average of all point's predictions.
    \item \textbf{Weighted Displacement with 6D orientation: } Instead of predicting the offset to the 4 goal end-effector points, this ablation predicts the 3D offset to the goal end-effector position, and the 6D orientation of the goal end-effector, from each scene point. The final prediction is the average of each point's prediction. 
    \item \textbf{Replacing low-level policy with a motion planner: } This ablation does not use a low-level policy for moving the robot end-effector. 
     We first predict a goal end-effector pose for grasping using the high-level policy and then use a motion planner to reach it. 
After grasping, we run the high-level policy again to predict a goal end-effector pose for opening the door. We compute the corresponding joint angles using inverse kinematics and use a joint PD controller to reach it. 
\end{itemize}
More details of these ablations can be found in Appendix~\ref{app:detail-ablation}. 
We compare to these ablations when training with 200 objects without camera randomizations. 
The results are shown in Table~\ref{tab:ablation}.  
We find that using a Point Transformer, not predicting the per-point weights, or predicting a per-point goal end-effector 6D orientation instead of per-point displacements to the goal end-effector points, all lead to worse performance, supporting the effectiveness of our design choices in \method{}. 
Predicting a per-point 6D orientation and averaging them leads to a large drop in performance because it is difficult to correctly compute the average of multiple 6D orientations; in \method{}, we avoid this by representing the goal end-effector pose as a collection of points and averaging the displacement to these points. 
Replacing the low-level policy with motion planning and an IK controller results in very poor performance.
We hypothesis it could be due to two reasons: 
1. The high-level policy may predict poses with minor collisions, and motion planning often fails due to the inability to find collision-free paths.
2. During door-opening, the joint PD controller takes the shortest joint-space path to the goal pose, ignoring necessary kinematic constraints (e.g., following an arc to open a revolute door), causing the gripper to slip off.
Experimentally, the motion planning failure rate is 17\%. Among successful grasps, 97\% of the failures are due to the gripper falling off the handle during opening
(possibly due to ignoring object kinematic constraints).
This shows the importance of using a learned low-level policy.   


\begin{table}[t]
\scriptsize
    \centering
        \centering
        \begin{tabular}{c|c|c}
           \toprule
           Method  & 
           \begin{tabular}[c]{@{}c@{}}Grasping\\Sucess Rate\end{tabular}
            & 
            \begin{tabular}[c]{@{}c@{}}Normalized \\Opening Performance\end{tabular}
            \\ \midrule
           \method{} (Ours) &  $\mathbf{0.88 \pm 0.01}$ &  $\mathbf{0.75 \pm 0.01}$ \\
           AO-Grasp & $0.11 \pm 0.0$ & $0.08 \pm 0.0$ \\ \midrule
           \method{} (Ours), After Grasping & - & $\mathbf{0.86 \pm 0.01}$ \\
           FlowBot3d - w/o Mask, After Grasping  & - & $0.2 \pm 0.01$ \\
           FlowBot3d - w/ Mask, After Grasping & - & $0.57 \pm 0.01$ \\
           \bottomrule
        \end{tabular}
    \caption{Comparison with prior articulated object manipulation methods.}
    \vspace{-6mm}
    \label{tab:prior_method}
\end{table}

\subsection{Comparison with Prior Articulated Object Manipulation Methods}
\label{sec:comparison with prior method}
We also compare our system with prior methods that aim to learn a single policy for generalizable articulated object manipulation. 
We compare to two state-of-the-art prior methods that focus on each stage of manipulating an articulated object:
\begin{itemize}
    \item \textbf{AO-Grasp}~\cite{morlans2024grasp}: this method focuses on the grasping of the articulated objects for downstream manipulation. It learns an Actionable Grasp Point Predictor that predicts the grasp-likelihood scores for each point in the point cloud, which is combined with pretrained ContactGrasp-Net~\cite{sundermeyer2021contact} to generate 6D grasps.
    \item \textbf{FlowBot3D}~\cite{eisner2022flowbot3d}: this method predicts the flow for each point on the articulated object, and moves the robot end-effector along the maximal flow direction to open the object. The original paper performs grasping by using a suction gripper to attach the robot end-effector to the maximal flow point on the object's surface. 
\end{itemize}
We compare \method{} with AO-Grasp in terms of \textbf{grasping success rate}, i.e., if the method generates a firm grasp of the object that enables downstream manipulation. As AO-Grasp only generates a 6D grasp pose, we use motion planning to move the robot end-effector to reach the grasping pose. 
Although AO-Grasp does not open the object, we still compare with it in terms of \textbf{normalized opening performance} in the following way:
After grasping, we assume access to ground-truth articulation information of the object and use our designed opening action (See Sec.~\ref{sec:demon_generation}) to open the object. Note such information is not available in the real world, and \method{} also does not use such information in the learned policy. 
We compare with FlowBot3D in terms of \textbf{normalized opening performance after grasping}: starting from the state where the robot gripper has already firmly grasped the handle of the object, how well does the method open the object. We used pre-trained checkpoints provided by the authors of AO-Grasp and FlowBot3D for the comparison. 

The results are shown in Table~\ref{tab:prior_method}, tested without camera randomizations. 
The grasping success rate of AO-Grasp is much lower than \method{}. We find that AO-Grasp often proposes grasps at the edges of the point cloud (e.g., the side wall of a drawer; see Appendix~\ref{app:detail-prior-articulated-object} for visuals). This likely stems from its training data, which includes many objects in a partially opened state where edge grasps are valid. However, in our test cases, objects are usually closed or not open enough for such grasps. Additionally, many detected edges are fake edges that result from partial observations from the camera rather than true graspable edges.
For FlowBot3D, we find its performance to be reasonable ($0.57$) when provided with the segmentation mask of the target link (door or drawer) to open, and much lower ($0.2$) without such masks. Note that \method{} does not use a segmentation mask for the target link. 
To form a fair comparison, we also evaluated our policy's performance after grasping. In such a case, the performance of \method{} further improved from $0.75$ to be $0.86$ (See  Table~\ref{tab:prior_method}), outperforming FlowBot3D by a large margin.

\subsection{Comparison of Different Low-level Policies}
We also performed experiments to test the performance of different low-level policy architectures (e.g., using a different diffusion head, or using a different action space).  
The detailed results and analysis can be found in Appendix~\ref{app:low-level-policy}. In summary, we do not observe huge performance differences (within 5\%) between these different methods.  
Our hypothesis is that the goal end-effector points provide a strong conditioning for the low-level policy; with such information as input, the differences in the policy architectures and action spaces may not matter too much.

\subsection{Additional Experiments and Evaluations}
We show some preliminary experiments in Appendix~\ref{app:more-task} that our hierarchical policy learning approach works on more manipulation tasks beyond articulated object manipulation.
We also study how robust the policy is to the orientation and position of the handles, with the results shown in Appendix~\ref{app:handle-augmentation}.


\begin{figure}[t]
    \centering
    \includegraphics[width=\linewidth]{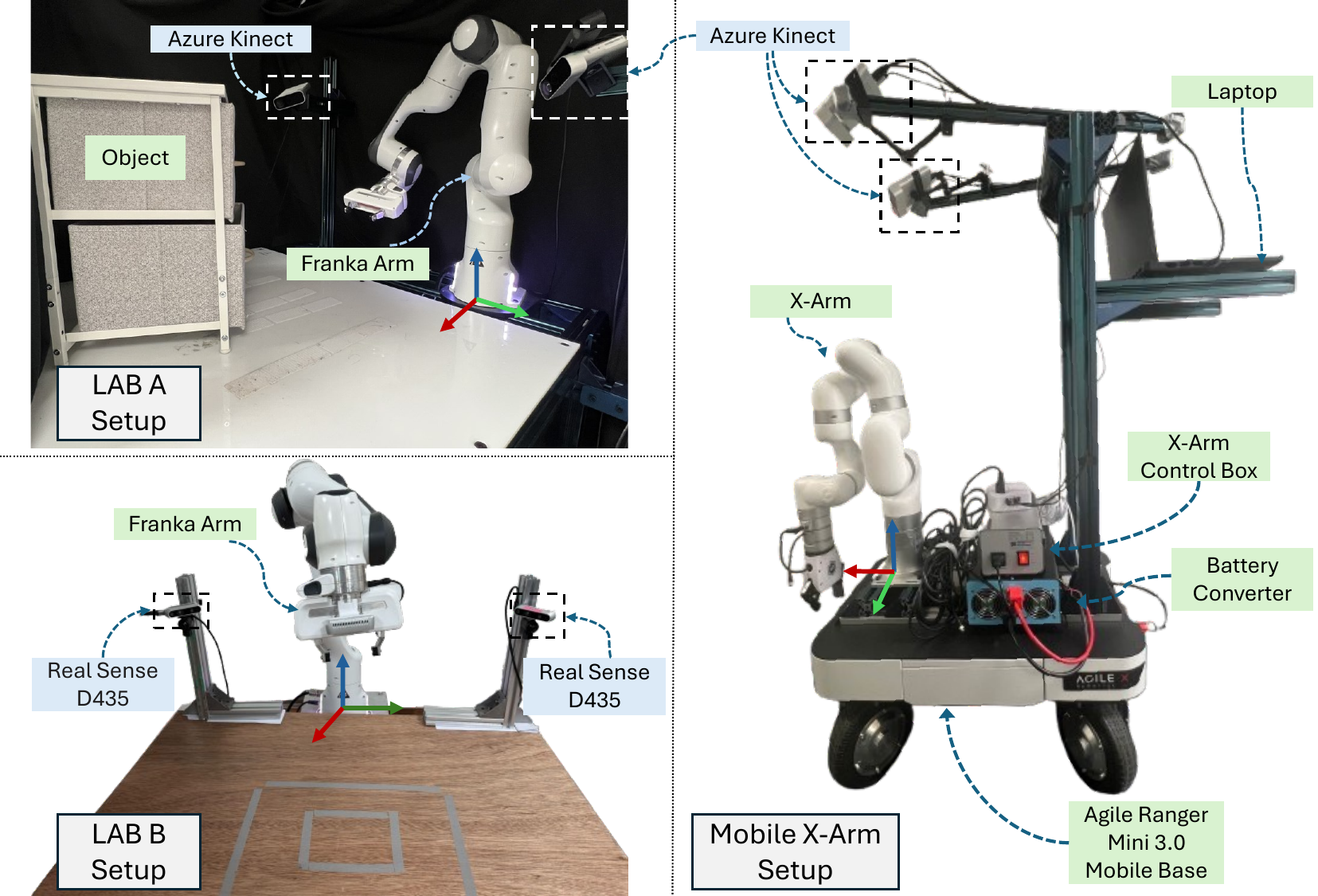}
    \vspace{-5mm}
    \caption{The three different real robot setups.}
    \vspace{-2mm}
    \label{fig:real_world_setup}
\end{figure}


\begin{figure}[t]
    \centering
     \includegraphics[width=\linewidth]{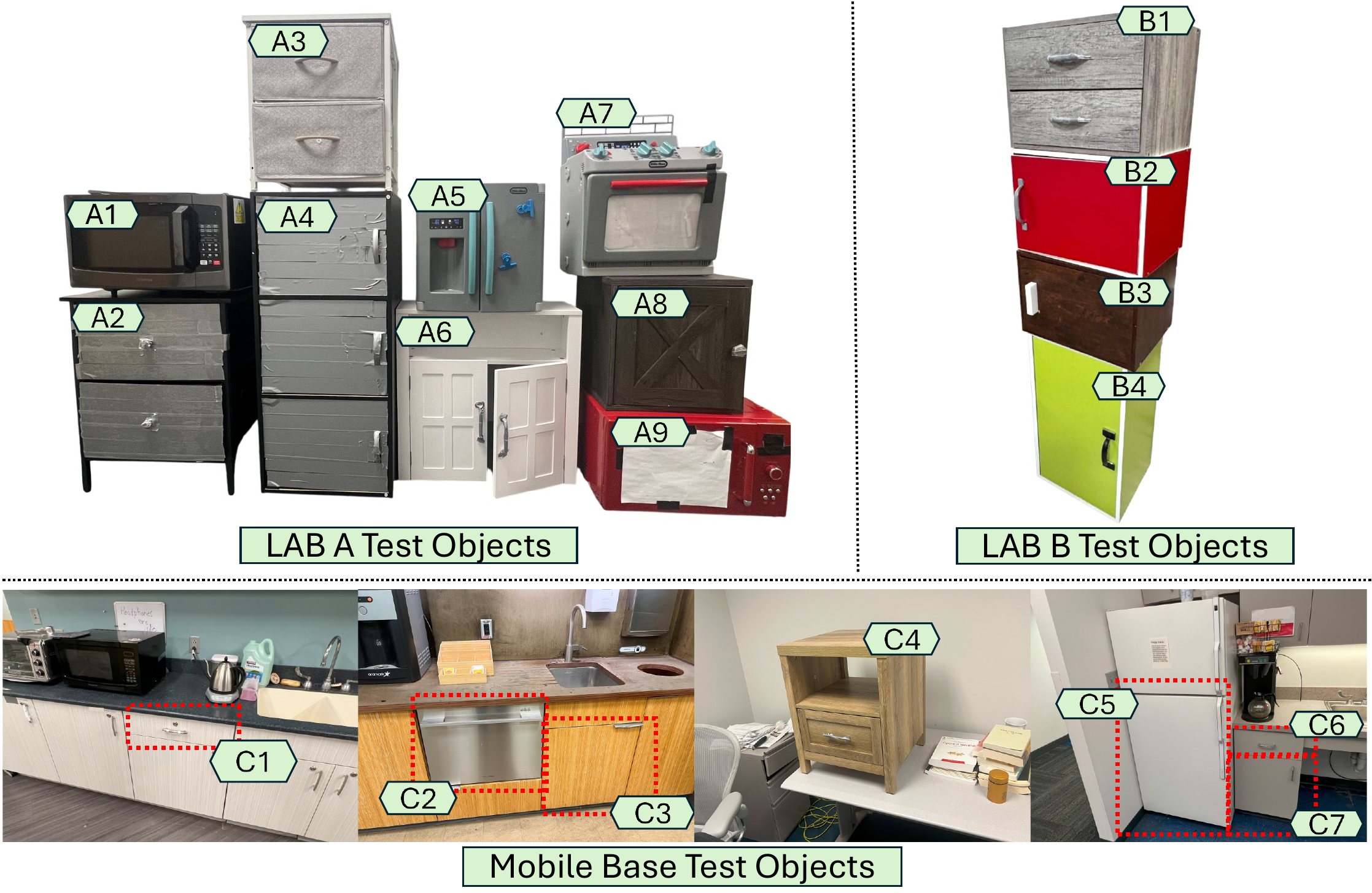}
    \vspace{-6mm}
    \caption{Real-world test objects for table-top and mobile-base experiments. }
    \vspace{-6mm}
    \label{fig:real_world_object}
\end{figure}

\begin{figure*}[t]
    \centering
    \includegraphics[width=\textwidth]{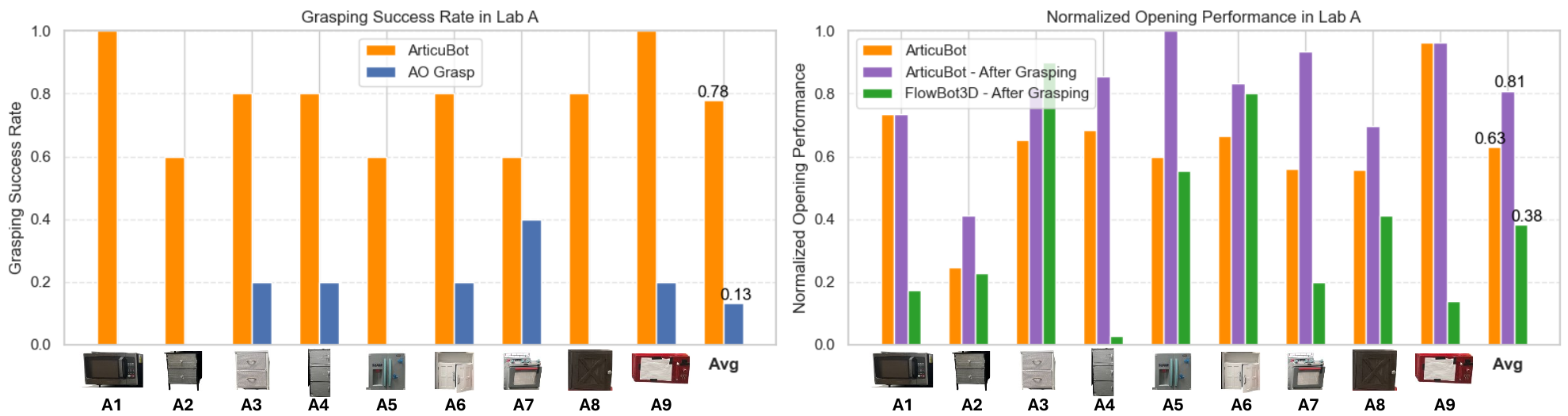}
    \vspace{-6mm}
    \caption{Comparison of \method{} with FlowBot3D and AO-Grasp on 9 test objects in Lab A with table-top Franka. We omit OpenVLA in the plot as it achieves a performance of 0. 
    }
    \label{fig:real_world_results}
    \vspace{-5mm}
\end{figure*}

\section{Real-world Experiments}
\label{sec:real-world-experiments}
\subsection{Setups}
We deploy our learned policies to three different real robot settings: table-top fixed Franka arms in two different labs, and an X-Arm on a moble base in real lounge and kitchens,  to test its robustness and generalization ability in the real world. 
We note that our policy in simulation is only trained on the Franka arm. The policy can transfer zero-shot cross embodiment to an X-Arm because the policy learns actions in the robotic arm's end-effector space (sub-goal end-effector pose and end-effector delta-transformations) instead of the joint space.

For robust sim2real transfer, we generate more demonstrations in simulation with augmentations on the point cloud observations to make the policy robust to noisy point clouds obtained from real-world depth sensors. 
Specifically, we add the following two augmentations to the depth map in simulation~\cite{dalal2024local}: the first is edge artifacts that models the noise along object edges, and the second is random holes in the depth map to model random depth pixel value loss in real-world depth cameras. Details of these augmentations can be found in the Appendix~\ref{app:point-cloud-augmentation}.
We also randomize the camera poses closer to where they are located in the real world. 
Combined with the non-augmented demonstrations, we generated in total 42.3k trajectories with 4.7M observation-action pairs, and trained a single weighted displacement model high-level policy on this dataset. 
We find the low-level policy to transfer well without needing to be trained on such point cloud augmentations. 
We detail the 3 different robot setups as below, visualized in Fig.~\ref{fig:real_world_setup}.

\noindent\textbf{Table-top Franka Panda Arm in Lab A: }
The first setup has a fixed-base table-top Franka Arm. The table has a length and width of 110 cm. The robot arm is located near one corner of the table. 
We use two Azure Kinect cameras, each mounted on one side of the robot looking at the center of the table, to get the point cloud of the objects. The robot is controlled via the Deoxys library~\cite{zhu2022viola} with a joint position controller, i.e., given a target pose, we first use a IK solver to obtain the target joint angle, and then use the joint position controller to reach that joint angle. We test 9 different articulated objects, including drawers, cabinets, microwaves, toy oven and toy fridges in this workspace (as shown in Fig.~\ref{fig:real_world_object}), all purchased from local stores and not seen during training.

\noindent\textbf{Table-top Franka Panda Arm in Lab B: }
We also deploy our policy in a different lab to more thoroughly test its robustness in a different setting. The table used in this lab has a width and length and width of 100cm and 80cm. The robot is placed at the center of one edge of the table. Two Intel RealSense D-435 RGBD cameras, one mounted on each side of the robot, are used to capture the objects' point clouds.
The robot is controlled using a end-effector position controller. 
Four different objects are tested in this workspace, shown in Fig.~\ref{fig:real_world_object}, all purchased from local stores and not seen during training.

\noindent\textbf{X-Arm on a mobile base: }
To test our policy in real lounges and kitchens, we additionally build a mobile manipulator, where we assemble an X-Arm onto a Ranger Mini 3.0 mobile base, following the design in~\citet{xiong2024adaptive}. We mount two Azure Kinects on manually built frames on the mobile base for capturing point cloud observations (see Fig.~\ref{fig:real_world_setup}).
We use the company-provided end-effector position control python API for controlling the X-Arm. 
We test this mobile X-Arm in 4 different kitchen, lounge and offices on 7 objects (See Fig.~\ref{fig:real_world_object}). 
The X-Arm and both Azure Kinects are connected to a Lenovo Legion Pro 7 Laptop. The laptop has a built-in NVIDIA GeForce RTX 4090 GPU, which is used for running the trained policies. 
The X-Arm, the Azure Kinects, and the laptop are all powered by the battery that comes with the Ranger Mini 3.0 mobile base, which gives a 48V DC output, and we use a bettery inverter to convert it to a standard 120V AC output to power these devices.

For both table-top settings, the objects are placed near the center of the table with some variations in the position and orientation, such that the robot arm can open it as much as possible within its joint limits. 
In Lab A, the Franka arm is randomly initialized at one of two fixed locations, one with the end-effector closer to the table, and the other with the end-effector higher in the air. 
In Lab B, the arm is randomly initialized such that the end-effector is 30 to 60 centimeters away from the object. 
For the mobile X-Arm, we manually tele-operate the base to be near the target object, and the base remains fixed when the X-Arm is opening the object. We randomly initialize the X-Arm at different joint angles. 

We perform camera-to-robot base calibration in all settings, and all point cloud observations are transformed into the robot's base frame. 
We use GroundingDino~\cite{liu2025grounding} and EfficientSAM~\cite{xiong2024efficientsam} to segment the object given a text of the object name, and obtain an object-only point cloud in the real world. 
For removing the robot from the point cloud, we first render a canonical robot point cloud from the robot urdf and mesh files, transform it to the current point cloud observation using the robot's current joint angles, and then project it to the 2D depth image using known camera extrinsics and intrinsics to obtain a robot mask. All pixels within the robot mask are removed. 
We perform additional radius and statistical outlier removing to remove some remaining outlier points from the noisy depth cameras. 
More details of the real-world perception pipeline can be found in Appendix~\ref{app:perception-pipeline}. 

We use the following evaluation metrics as in simulation.
    \textbf{Grasping Success Rate}: We manually check if the robot gripper has a firm grasp of the object. 
    \textbf{Normalized Opening Performance}: The opened distance of the object normalized by the maximal achievable opening distance of the object, subject to the workspace and robot joint limit constraints. 

We compare to the following baselines with the table-top Franka Arm in lab A, on 9 test objects. \textbf{OpenVLA}: This is an open-sourced robotic foundation policy trained on Internet scale of real-world datasets. It takes a language instruction as input and output robot actions. A small portion of the datasets contain articulated object manipulation tasks. We also compare to \textbf{AO-Grasp} and \textbf{FlowBot3D} as described in Sec.~\ref{sec:comparison with prior method}. We compare with AO-Grasp in terms of grasping success rate: we use our learned low-level policy to reach the grasping pose generated by AO-Grasp and manually check if the grasping is successful. We compare with FlowBot3D in terms of normalized opening performance: we first manually move the end-effector to grasp the handle of the object, and then apply FlowBot3D to open the object. We do not input the optional segmentation mask for the target link to open for FlowBot3D, as such masks are not readily available in the real world, and \method{} also does not use such masks.
For table-top Franka arms, we run each method on each object for five trials and report the mean performance. 
For mobile X-Arm, we run \method{} for three trials on each object. 


\begin{table}[t]
\scriptsize
    \centering
        \centering
        \begin{tabular}{c|c|c}
           \toprule
           Robot Test Settings  & 
           \begin{tabular}[c]{@{}c@{}}Grasping\\Sucess Rate\end{tabular}
            & 
            \begin{tabular}[c]{@{}c@{}}Normalized \\Opening Performance\end{tabular}
            \\ \midrule
           \method{} - Tabletop Franka Lab A &  0.78 & 0.63 \\
           \method{} - Tabletop Franka Lab B & 0.85 & 0.59 \\
           \method{} - Mobile X-Arm & 0.90 & 0.54 \\
           \bottomrule
        \end{tabular}
    \caption{Performance of \method{} under all three robot setups.}
    \vspace{-7mm}
    \label{tab:real-world-performance}
\end{table}

\begin{figure*}[t]
    \centering
    \includegraphics[width=\textwidth]{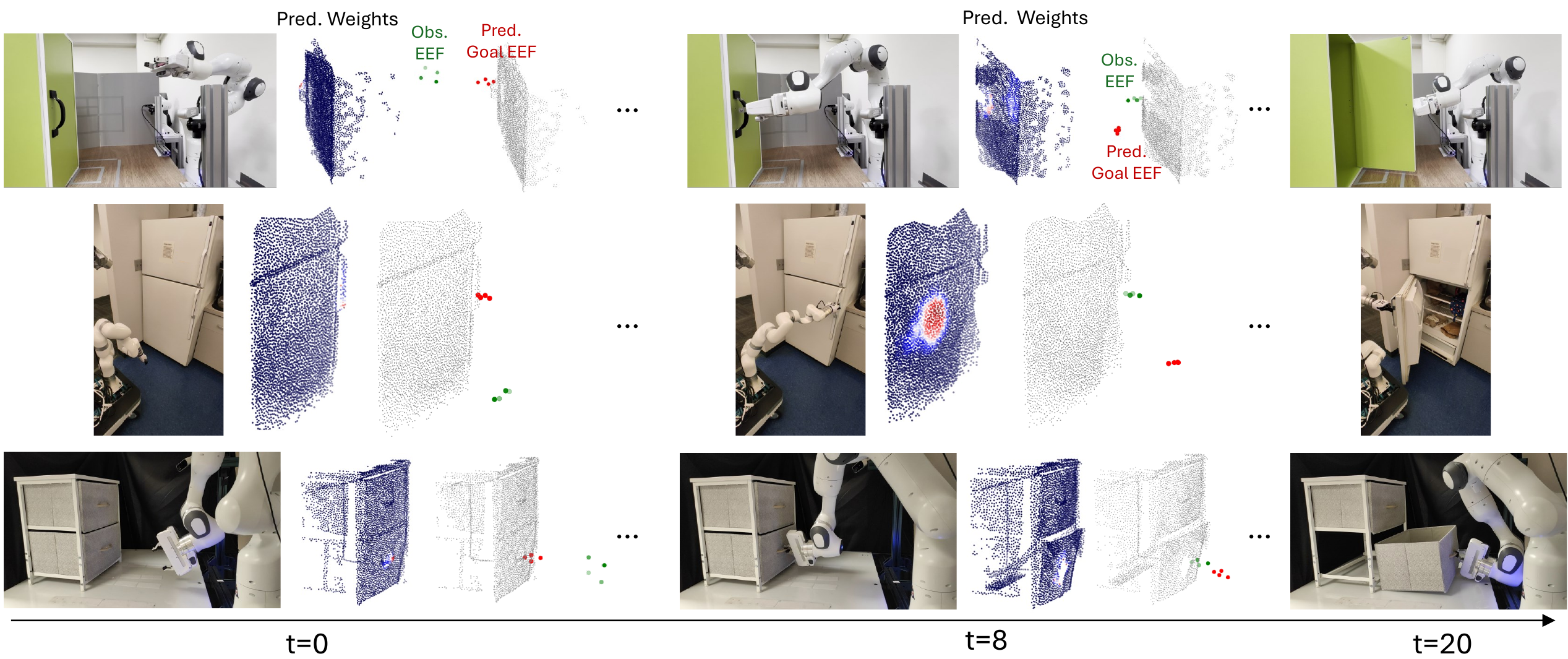}
    \vspace{-6mm}
    \caption{Visualizations of the high-level policy's predictions (per-point weights and goal end-effector points) in three of the real-world test cases. 
    The green points represent the observed current end-effector points, and the red points represent the predicted goal end-effector points.
    }
    \label{fig:real_world_visuals}
    \vspace{-6mm}
\end{figure*}

\subsection{Table-Top Franka Arm Results}
The results for all test objects and compared methods in lab A are shown in Fig.~\ref{fig:real_world_results}; the results 
of \method{} in lab B are shown in Table~\ref{tab:real-world-performance}. 
\method{} achieves an average grasping success rate of $0.78$ and $0.85$, and a normalized opening performance of $0.63$ and $0.59$ in Lab A and B, respectively, showing that it can generalize to open diverse real articulated objects with varying geometries, shapes and articulation types across both lab settings. 
We note that some of the test objects are quite challenging and require very precise manipulation, e.g., the knobs of object A2 and A8 are very small, with a diameter of only 2 cm, but \method{} can still precisely grasp and open it. 
As in simulation, we find the grasping success rate of AO-grasp to be low, where it tends to grasp at the ``fake'' edge in the point cloud due to partial observations, which are not actually graspable (See Appendix~\ref{app:detail-prior-articulated-object} for visuals of the grasps produced by AO-Grasp). 
FlowBot3D achieves a reasonable normalized opening performance of $0.38$, starting from the state where the robot gripper already grasps the object. The performance is still lower than \method{}, even though \method{} performs the additional grasping step. 
The major failure case for FlowBot3D is that the predicted flow is in the wrong direction, e.g., it predicts upwards flows for opening a microwave (See Appendix~\ref{app:detail-prior-articulated-object} for visuals of the flows). 
If we compute the normalized opening performance for \method{} only in cases where the grasp is successful (i.e., the same starting conditions as FlowBot3D), the performance of \method{} further improves to $0.81$ and outperforms FlowBot3D by a large margin (See Fig.~\ref{fig:real_world_results}). 
We also find OpenVLA fails to grasp or open any test objects, resulting in a grasping success rate and normalized opening performance of 0. This is likely due to its training data lacking sufficient demonstrations of articulated object manipulation, making generalization to our test cases difficult.

Fig.~\ref{fig:real_world_visuals} (zoom-in for better views) visualizes \method{}'s predictions on some of the real-world test objects. As shown, the learned per-point weights from the weighted displacement model concentrates on the handle of the object before grasping, and the predicted grasping end-effector pose is quite accurate on the handle, even though for most of the objects the handles are just a very small portion of the point cloud. We note that there is no explicit supervision for the model to learn to assign high weights to the handle; this is automatically learned by just minimizing the imitation learning loss. 
After grasping, the weights are more randomly distributed across the objects; but as shown in Fig.~\ref{fig:real_world_visuals}, \method{} generalizes to predict different opening end-effector poses for objects with different articulations (left opening revolute joints, right opening revolute joints, pulling out prismatic joints). 
We do notice a drop in performance compared to the results in simulation. 
We believe this is mostly due to the noisier point cloud observation from the depth sensors in the real world (e.g., see the noisy point cloud for the green cabinet obtained from the RealSense cameras in Fig.~\ref{fig:real_world_visuals}).
See Fig.~\ref{fig:draw} for screenshots of the opening trajectories for more objects (and Appendix~\ref{app:real_world_visual} for trajectories of all objects); please refer to the supplementary materials for videos. 
Common failure cases for table-top experiments include: 1. The robot arm runs into joint limits while opening the object, due to the limited space of the robot workstation. 2. Wrong end-effector pose predictions for grasping the handle, which we find to happen more for objects with small handles, e.g., A2. 
Appendix~\ref{app:failure-case} provides visualizations of some of the failure cases.

\subsection{Mobile X-Arm Results}
Table~\ref{tab:real-world-performance} shows the results with the mobile X-Arm. 
As shown, \method{} achieves a grasping success rate of $0.9$ and normalized opening performance of $0.54$, showing it can generalize to drawers, cabinets, and fridges in real kitchens and lounges. 
See Fig.~\ref{fig:real_world_visuals} for a visualization of the high-level policy predictions from \method{} on a real-world fridge.  
See Fig.~\ref{fig:draw} for screenshots of some of the successful opening trajectories with the mobile X-Arm (and Appendix~\ref{app:real_world_visual} for trajectories of all objects); please refer to the supplementary materials for videos. 
Some of the tested real-world objects are quite challenging, for example, cabinet C3 has a very small handle that protrudes only 2 cm from the surface, but \method{} is still capable of precisely grasping and opening it.

We do notice a drop in the normalized opening performance compared to the table-top Franka experiments. 
In our early experiments, we find that Unlike the Franka Arm, the X-Arm lacks impedance control and force sensing.  This requires a more precise prediction for the opening end-effector pose. Small prediction errors, such as turning too sharply when opening a revolute door, would result in excessive force for the X-Arm and causes it to stop for motor protection, which is a common failure case in this setting. 
To partially mitigate this issue, we use the Fast-UMI~\cite{wu2024fast} gripper in latter experiments, whose soft, compliant design enhances safety and partially helps prevent the arm from stopping due to excessive force.
Another sim2real gap is that, in simulation, objects are isolated by itself; in real kitchens and lounges, objects are usually surrounded by other objects, which occlude its side and top, and only the front side is observable.  This additional occlusion might have caused issues for the policy to transfer as well.
Finally, the articulation of some objects are inherently ambiguous to judge from a single point cloud observation, e.g., although many dishwashers have a revolute door that open downwards, the dishwasher C2 in Fig.~\ref{fig:real_world_object} has a prismatic joint that needs to be pulled out horizontally. \method{} tends to make more mistakes on such ambiguous objects. See Appendix~\ref{app:failure-case} for visualizations of some of the failure cases of \method{}, and some basic failure recovery abilities of \method{}.  



\section{Limitations}
Our system currently has the following limitations:
1) Our weighted displacement policy does not handle multi-modal outputs since it is trained with a regression loss (Eq.~\eqref{eq:high-level-loss}). This may create issues for it when working with cabinets with multiple doors and opening only a specific one is desired.
2) The current system does not support opening a user-specified door on a multi-door object, as the policy is not trained to be conditioned on any user input. 
Although our training data includes multi-door objects,  demonstrations are generated for opening the closest door to the initial pose of the robot. The policy learns to open the closest door implicitly, rather than opening a user specified door. 
3) The policy uses point cloud observations. Although this simplifies sim2real transfer, existing depth sensors in the real world do not work well for transparent or reflective objects, and thus our policy would also fail on such objects in the current form. 
4) As mentioned above, the X-Arm itself does not support force sensing and impedance control, which requires more precise policy predictions to avoid excessive force. 
We think adding a force-torque sensor on the X-Arm to enable impedance control could help alleviate this issue; fine-tuning the policy in the real-world via reinforcement learning or a few demonstrations for more precise sub-goal end-effector pose predictions could also help. 
5) Objects in real kitchens and lounges are usually occluded by neighboring objects, and we believe that adding this type of occlusion could further improve the sim2real performance of the policy. 
6) \method{} does not use interaction history during the manipulation process. We think that incorporating interaction history with current visual observations could further improve performance, especially for objects whose articulation are ambiguous to judge just from visual observations. 
We leave addressing these limitations as important future work. 

\section{Conclusion} 
This paper presents \method, a robotics system powered by a single learned policy that is able to open diverse categories of unseen articulated objects in the real world.   
\method{} consists of three parts: generating a large number of demonstrations in simulation, distilling all generated demonstrations into a point cloud-based neural policy via imitation learning, and performing zero-shot sim2real transfer.
Using sampling-based grasping and motion planning, \method{}'s demonstration generalization pipeline is fast and effective, generating a total of 42.3k demonstrations over 322 training articulated objects.
For policy learning, \method{} uses a novel hierarchical policy representation, in which the high-level policy learns the sub-goal for the end-effector, and the low-level policy learns how to move the end-effector conditioned on the predicted goal. 
A novel weighted displacement model is used for the high-level policy that grounds the prediction into the existing 3D structure of the scene, outperforming alternative policy representations. 
Our learned policy can zero-shot transfer to three different real robot settings: a fixed table-top Franka arm across two different labs, and an X-Arm on a mobile base, opening multiple unseen articulated objects across two labs, real lounges, and kitchens.   

\section*{Acknowledgments}
This material is based upon work supported by the Toyota Research Institute,  National Science Foundation under NSF CAREER Grant No. IIS-2046491, and NIST under Grant No. 70NANB24H314. Any opinions, findings, and
 conclusions or recommendations expressed in this material are
 those of the author(s) and do not necessarily reflect the views
 of Toyota Research Institute, National Science Foundation, or NIST.


\bibliographystyle{plainnat}
\bibliography{references}

\begin{thebibliography}{67}
\providecommand{\natexlab}[1]{#1}
\providecommand{\url}[1]{\texttt{#1}}
\expandafter\ifx\csname urlstyle\endcsname\relax
  \providecommand{\doi}[1]{doi: #1}\else
  \providecommand{\doi}{doi: \begingroup \urlstyle{rm}\Url}\fi

\bibitem[Akkaya et~al.(2019)Akkaya, Andrychowicz, Chociej, Litwin, McGrew, Petron, Paino, Plappert, Powell, Ribas, et~al.]{akkaya2019solving}
Ilge Akkaya, Marcin Andrychowicz, Maciek Chociej, Mateusz Litwin, Bob McGrew, Arthur Petron, Alex Paino, Matthias Plappert, Glenn Powell, Raphael Ribas, et~al.
\newblock Solving rubik's cube with a robot hand.
\newblock \emph{arXiv preprint arXiv:1910.07113}, 2019.

\bibitem[Brohan et~al.(2022)Brohan, Brown, Carbajal, Chebotar, Dabis, Finn, Gopalakrishnan, Hausman, Herzog, Hsu, et~al.]{brohan2022rt}
Anthony Brohan, Noah Brown, Justice Carbajal, Yevgen Chebotar, Joseph Dabis, Chelsea Finn, Keerthana Gopalakrishnan, Karol Hausman, Alex Herzog, Jasmine Hsu, et~al.
\newblock Rt-1: Robotics transformer for real-world control at scale.
\newblock \emph{arXiv preprint arXiv:2212.06817}, 2022.

\bibitem[Brohan et~al.(2023)Brohan, Brown, Carbajal, Chebotar, Chen, Choromanski, Ding, Driess, Dubey, Finn, et~al.]{brohan2023rt}
Anthony Brohan, Noah Brown, Justice Carbajal, Yevgen Chebotar, Xi~Chen, Krzysztof Choromanski, Tianli Ding, Danny Driess, Avinava Dubey, Chelsea Finn, et~al.
\newblock Rt-2: Vision-language-action models transfer web knowledge to robotic control.
\newblock \emph{arXiv preprint arXiv:2307.15818}, 2023.

\bibitem[Chen et~al.(2023)Chen, Tippur, Wu, Kumar, Adelson, and Agrawal]{chen2023visual}
Tao Chen, Megha Tippur, Siyang Wu, Vikash Kumar, Edward Adelson, and Pulkit Agrawal.
\newblock Visual dexterity: In-hand reorientation of novel and complex object shapes.
\newblock \emph{Science Robotics}, 8\penalty0 (84):\penalty0 eadc9244, 2023.

\bibitem[Chen et~al.(2024)Chen, Walsman, Memmel, Mo, Fang, Vemuri, Wu, Fox, and Gupta]{chen2024urdformer}
Zoey Chen, Aaron Walsman, Marius Memmel, Kaichun Mo, Alex Fang, Karthikeya Vemuri, Alan Wu, Dieter Fox, and Abhishek Gupta.
\newblock Urdformer: A pipeline for constructing articulated simulation environments from real-world images.
\newblock \emph{arXiv preprint arXiv:2405.11656}, 2024.

\bibitem[Cheng et~al.(2024)Cheng, Shi, Agarwal, and Pathak]{cheng2024extreme}
Xuxin Cheng, Kexin Shi, Ananye Agarwal, and Deepak Pathak.
\newblock Extreme parkour with legged robots.
\newblock In \emph{2024 IEEE International Conference on Robotics and Automation (ICRA)}, pages 11443--11450. IEEE, 2024.

\bibitem[Coumans and Bai(2016--2021)]{coumans2021}
Erwin Coumans and Yunfei Bai.
\newblock Pybullet, a python module for physics simulation for games, robotics and machine learning.
\newblock \url{http://pybullet.org}, 2016--2021.

\bibitem[Dalal et~al.(2024{\natexlab{a}})Dalal, Liu, Talbott, Chen, Pathak, Zhang, and Salakhutdinov]{dalal2024local}
Murtaza Dalal, Min Liu, Walter Talbott, Chen Chen, Deepak Pathak, Jian Zhang, and Ruslan Salakhutdinov.
\newblock Local policies enable zero-shot long-horizon manipulation.
\newblock \emph{arXiv preprint arXiv:2410.22332}, 2024{\natexlab{a}}.

\bibitem[Dalal et~al.(2024{\natexlab{b}})Dalal, Yang, Mendonca, Khaky, Salakhutdinov, and Pathak]{dalal2024neural}
Murtaza Dalal, Jiahui Yang, Russell Mendonca, Youssef Khaky, Ruslan Salakhutdinov, and Deepak Pathak.
\newblock Neural mp: A generalist neural motion planner.
\newblock \emph{arXiv preprint arXiv:2409.05864}, 2024{\natexlab{b}}.

\bibitem[Eisner et~al.(2022)Eisner, Zhang, and Held]{eisner2022flowbot3d}
Ben Eisner, Harry Zhang, and David Held.
\newblock Flowbot3d: Learning 3d articulation flow to manipulate articulated objects.
\newblock \emph{arXiv preprint arXiv:2205.04382}, 2022.

\bibitem[Etukuru et~al.(2024)Etukuru, Naka, Hu, Lee, Mehu, Edsinger, Paxton, Chintala, Pinto, and Shafiullah]{etukuru2024robot}
Haritheja Etukuru, Norihito Naka, Zijin Hu, Seungjae Lee, Julian Mehu, Aaron Edsinger, Chris Paxton, Soumith Chintala, Lerrel Pinto, and Nur Muhammad~Mahi Shafiullah.
\newblock Robot utility models: General policies for zero-shot deployment in new environments.
\newblock \emph{arXiv preprint arXiv:2409.05865}, 2024.

\bibitem[Fang et~al.(2023)Fang, Wang, Fang, Gou, Liu, Yan, Liu, Xie, and Lu]{fang2023anygrasp}
Hao-Shu Fang, Chenxi Wang, Hongjie Fang, Minghao Gou, Jirong Liu, Hengxu Yan, Wenhai Liu, Yichen Xie, and Cewu Lu.
\newblock Anygrasp: Robust and efficient grasp perception in spatial and temporal domains.
\newblock \emph{IEEE Transactions on Robotics}, 2023.

\bibitem[Fishman et~al.(2023)Fishman, Murali, Eppner, Peele, Boots, and Fox]{fishman2023motion}
Adam Fishman, Adithyavairavan Murali, Clemens Eppner, Bryan Peele, Byron Boots, and Dieter Fox.
\newblock Motion policy networks.
\newblock In \emph{Conference on Robot Learning}, pages 967--977. PMLR, 2023.

\bibitem[Gammell et~al.(2015)Gammell, Srinivasa, and Barfoot]{gammell2015batch}
Jonathan~D Gammell, Siddhartha~S Srinivasa, and Timothy~D Barfoot.
\newblock Batch informed trees (bit*): Sampling-based optimal planning via the heuristically guided search of implicit random geometric graphs.
\newblock In \emph{2015 IEEE international conference on robotics and automation (ICRA)}, pages 3067--3074. IEEE, 2015.

\bibitem[Gupta et~al.(2024)Gupta, Zhang, Sathua, and Gupta]{gupta2024opening}
Arjun Gupta, Michelle Zhang, Rishik Sathua, and Saurabh Gupta.
\newblock Opening cabinets and drawers in the real world using a commodity mobile manipulator.
\newblock \emph{arXiv preprint arXiv:2402.17767}, 2024.

\bibitem[Ha and Song(2022)]{ha2022flingbot}
Huy Ha and Shuran Song.
\newblock Flingbot: The unreasonable effectiveness of dynamic manipulation for cloth unfolding.
\newblock In \emph{Conference on Robot Learning}, pages 24--33. PMLR, 2022.

\bibitem[Hauser and Ng-Thow-Hing(2010)]{hauser2010fast}
Kris Hauser and Victor Ng-Thow-Hing.
\newblock Fast smoothing of manipulator trajectories using optimal bounded-acceleration shortcuts.
\newblock In \emph{2010 IEEE international conference on robotics and automation}, pages 2493--2498. IEEE, 2010.

\bibitem[Hua et~al.(2024)Hua, Liu, Macaluso, Lin, Zhang, Xu, and Wang]{hua2024gensim2}
Pu~Hua, Minghuan Liu, Annabella Macaluso, Yunfeng Lin, Weinan Zhang, Huazhe Xu, and Lirui Wang.
\newblock Gensim2: Scaling robot data generation with multi-modal and reasoning llms.
\newblock \emph{arXiv preprint arXiv:2410.03645}, 2024.

\bibitem[Jain et~al.(2021)Jain, Lioutikov, Chuck, and Niekum]{jain2021screwnet}
Ajinkya Jain, Rudolf Lioutikov, Caleb Chuck, and Scott Niekum.
\newblock Screwnet: Category-independent articulation model estimation from depth images using screw theory.
\newblock In \emph{2021 IEEE International Conference on Robotics and Automation (ICRA)}, pages 13670--13677. IEEE, 2021.

\bibitem[Jain et~al.(2022)Jain, Giguere, Lioutikov, and Niekum]{pmlr-v164-jain22a}
Ajinkya Jain, Stephen Giguere, Rudolf Lioutikov, and Scott Niekum.
\newblock Distributional depth-based estimation of object articulation models.
\newblock In Aleksandra Faust, David Hsu, and Gerhard Neumann, editors, \emph{Proceedings of the 5th Conference on Robot Learning}, volume 164 of \emph{Proceedings of Machine Learning Research}, pages 1611--1621. PMLR, 08--11 Nov 2022.
\newblock URL \url{https://proceedings.mlr.press/v164/jain22a.html}.

\bibitem[Jiang et~al.(2022)Jiang, Hsu, and Zhu]{jiang2022ditto}
Zhenyu Jiang, Cheng-Chun Hsu, and Yuke Zhu.
\newblock Ditto: Building digital twins of articulated objects from interaction.
\newblock In \emph{Proceedings of the IEEE/CVF Conference on Computer Vision and Pattern Recognition}, pages 5616--5626, 2022.

\bibitem[Karaman and Frazzoli(2011)]{karaman2011sampling}
Sertac Karaman and Emilio Frazzoli.
\newblock Sampling-based algorithms for optimal motion planning.
\newblock \emph{The international journal of robotics research}, 30\penalty0 (7):\penalty0 846--894, 2011.

\bibitem[Ke et~al.(2024)Ke, Gkanatsios, and Fragkiadaki]{ke20243d}
Tsung-Wei Ke, Nikolaos Gkanatsios, and Katerina Fragkiadaki.
\newblock 3d diffuser actor: Policy diffusion with 3d scene representations.
\newblock \emph{arXiv preprint arXiv:2402.10885}, 2024.

\bibitem[Kim et~al.(2024)Kim, Pertsch, Karamcheti, Xiao, Balakrishna, Nair, Rafailov, Foster, Lam, Sanketi, et~al.]{kim2024openvla}
Moo~Jin Kim, Karl Pertsch, Siddharth Karamcheti, Ted Xiao, Ashwin Balakrishna, Suraj Nair, Rafael Rafailov, Ethan Foster, Grace Lam, Pannag Sanketi, et~al.
\newblock Openvla: An open-source vision-language-action model.
\newblock \emph{arXiv preprint arXiv:2406.09246}, 2024.

\bibitem[Kumar et~al.(2021)Kumar, Fu, Pathak, and Malik]{kumar2021rma}
Ashish Kumar, Zipeng Fu, Deepak Pathak, and Jitendra Malik.
\newblock Rma: Rapid motor adaptation for legged robots.
\newblock \emph{arXiv preprint arXiv:2107.04034}, 2021.

\bibitem[Lee et~al.(2020)Lee, Hwangbo, Wellhausen, Koltun, and Hutter]{lee2020learning}
Joonho Lee, Jemin Hwangbo, Lorenz Wellhausen, Vladlen Koltun, and Marco Hutter.
\newblock Learning quadrupedal locomotion over challenging terrain.
\newblock \emph{Science robotics}, 5\penalty0 (47):\penalty0 eabc5986, 2020.

\bibitem[Lin et~al.(2024)Lin, Hu, Sheng, Wen, You, and Gao]{lin2024data}
Fanqi Lin, Yingdong Hu, Pingyue Sheng, Chuan Wen, Jiacheng You, and Yang Gao.
\newblock Data scaling laws in imitation learning for robotic manipulation.
\newblock \emph{arXiv preprint arXiv:2410.18647}, 2024.

\bibitem[Liu et~al.(2023)Liu, Mahdavi-Amiri, and Savva]{liu2023paris}
Jiayi Liu, Ali Mahdavi-Amiri, and Manolis Savva.
\newblock Paris: Part-level reconstruction and motion analysis for articulated objects.
\newblock In \emph{Proceedings of the IEEE/CVF International Conference on Computer Vision}, pages 352--363, 2023.

\bibitem[Liu et~al.(2025)Liu, Zeng, Ren, Li, Zhang, Yang, Jiang, Li, Yang, Su, et~al.]{liu2025grounding}
Shilong Liu, Zhaoyang Zeng, Tianhe Ren, Feng Li, Hao Zhang, Jie Yang, Qing Jiang, Chunyuan Li, Jianwei Yang, Hang Su, et~al.
\newblock Grounding dino: Marrying dino with grounded pre-training for open-set object detection.
\newblock In \emph{European Conference on Computer Vision}, pages 38--55. Springer, 2025.

\bibitem[Mandi et~al.(2024)Mandi, Weng, Bauer, and Song]{mandi2024real2code}
Zhao Mandi, Yijia Weng, Dominik Bauer, and Shuran Song.
\newblock Real2code: Reconstruct articulated objects via code generation.
\newblock \emph{arXiv preprint arXiv:2406.08474}, 2024.

\bibitem[Mo et~al.(2021)Mo, Guibas, Mukadam, Gupta, and Tulsiani]{mo2021where2act}
Kaichun Mo, Leonidas~J Guibas, Mustafa Mukadam, Abhinav Gupta, and Shubham Tulsiani.
\newblock Where2act: From pixels to actions for articulated 3d objects.
\newblock In \emph{Proceedings of the IEEE/CVF International Conference on Computer Vision}, pages 6813--6823, 2021.

\bibitem[Morlans et~al.(2024)Morlans, Chen, Weng, Yi, Huang, Heppert, Zhou, Guibas, and Bohg]{morlans2024grasp}
Carlota~Par{\'e}s Morlans, Claire Chen, Yijia Weng, Michelle Yi, Yuying Huang, Nick Heppert, Linqi Zhou, Leonidas Guibas, and Jeannette Bohg.
\newblock Ao-grasp: Articulated object grasp generation.
\newblock In \emph{2024 IEEE/RSJ International Conference on Intelligent Robots and Systems (IROS)}, pages 13096--13103. IEEE, 2024.

\bibitem[O'Neill et~al.(2023)O'Neill, Rehman, Gupta, Maddukuri, Gupta, Padalkar, Lee, Pooley, Gupta, Mandlekar, et~al.]{o2023open}
Abby O'Neill, Abdul Rehman, Abhinav Gupta, Abhiram Maddukuri, Abhishek Gupta, Abhishek Padalkar, Abraham Lee, Acorn Pooley, Agrim Gupta, Ajay Mandlekar, et~al.
\newblock Open x-embodiment: Robotic learning datasets and rt-x models.
\newblock \emph{arXiv preprint arXiv:2310.08864}, 2023.

\bibitem[Perez et~al.(2018)Perez, Strub, De~Vries, Dumoulin, and Courville]{perez2018film}
Ethan Perez, Florian Strub, Harm De~Vries, Vincent Dumoulin, and Aaron Courville.
\newblock Film: Visual reasoning with a general conditioning layer.
\newblock In \emph{Proceedings of the AAAI conference on artificial intelligence}, volume~32, 2018.

\bibitem[Qi et~al.(2017{\natexlab{a}})Qi, Su, Mo, and Guibas]{qi2017pointnet}
Charles~R Qi, Hao Su, Kaichun Mo, and Leonidas~J Guibas.
\newblock Pointnet: Deep learning on point sets for 3d classification and segmentation.
\newblock In \emph{Proceedings of the IEEE conference on computer vision and pattern recognition}, pages 652--660, 2017{\natexlab{a}}.

\bibitem[Qi et~al.(2017{\natexlab{b}})Qi, Yi, Su, and Guibas]{qi2017pointnet++}
Charles~Ruizhongtai Qi, Li~Yi, Hao Su, and Leonidas~J Guibas.
\newblock Pointnet++: Deep hierarchical feature learning on point sets in a metric space.
\newblock \emph{Advances in neural information processing systems}, 30, 2017{\natexlab{b}}.

\bibitem[Qi et~al.(2023)Qi, Yi, Suresh, Lambeta, Ma, Calandra, and Malik]{qi2023general}
Haozhi Qi, Brent Yi, Sudharshan Suresh, Mike Lambeta, Yi~Ma, Roberto Calandra, and Jitendra Malik.
\newblock General in-hand object rotation with vision and touch.
\newblock In \emph{Conference on Robot Learning}, pages 2549--2564. PMLR, 2023.

\bibitem[Radosavovic et~al.(2024)Radosavovic, Xiao, Zhang, Darrell, Malik, and Sreenath]{radosavovic2024real}
Ilija Radosavovic, Tete Xiao, Bike Zhang, Trevor Darrell, Jitendra Malik, and Koushil Sreenath.
\newblock Real-world humanoid locomotion with reinforcement learning.
\newblock \emph{Science Robotics}, 9\penalty0 (89):\penalty0 eadi9579, 2024.

\bibitem[Ren et~al.(2024)Ren, Liu, Zeng, Lin, Li, Cao, Chen, Huang, Chen, Yan, et~al.]{ren2024grounded}
Tianhe Ren, Shilong Liu, Ailing Zeng, Jing Lin, Kunchang Li, He~Cao, Jiayu Chen, Xinyu Huang, Yukang Chen, Feng Yan, et~al.
\newblock Grounded sam: Assembling open-world models for diverse visual tasks.
\newblock \emph{arXiv preprint arXiv:2401.14159}, 2024.

\bibitem[Seita et~al.(2023)Seita, Wang, Shetty, Li, Erickson, and Held]{seita2023toolflownet}
Daniel Seita, Yufei Wang, Sarthak~J Shetty, Edward~Yao Li, Zackory Erickson, and David Held.
\newblock Toolflownet: Robotic manipulation with tools via predicting tool flow from point clouds.
\newblock In \emph{Conference on Robot Learning}, pages 1038--1049. PMLR, 2023.

\bibitem[Shridhar et~al.(2023)Shridhar, Manuelli, and Fox]{shridhar2023perceiver}
Mohit Shridhar, Lucas Manuelli, and Dieter Fox.
\newblock Perceiver-actor: A multi-task transformer for robotic manipulation.
\newblock In \emph{Conference on Robot Learning}, pages 785--799. PMLR, 2023.

\bibitem[Song et~al.(2020)Song, Meng, and Ermon]{song2020denoising}
Jiaming Song, Chenlin Meng, and Stefano Ermon.
\newblock Denoising diffusion implicit models.
\newblock \emph{arXiv preprint arXiv:2010.02502}, 2020.

\bibitem[Strub and Gammell(2020)]{strub2020advanced}
Marlin~P Strub and Jonathan~D Gammell.
\newblock Advanced bit*(abit*): Sampling-based planning with advanced graph-search techniques.
\newblock In \emph{2020 IEEE International Conference on Robotics and Automation (ICRA)}, pages 130--136. IEEE, 2020.

\bibitem[Su et~al.(2024)Su, Ahmed, Lu, Pan, Bo, and Liu]{su2024roformer}
Jianlin Su, Murtadha Ahmed, Yu~Lu, Shengfeng Pan, Wen Bo, and Yunfeng Liu.
\newblock Roformer: Enhanced transformer with rotary position embedding.
\newblock \emph{Neurocomputing}, 568:\penalty0 127063, 2024.

\bibitem[Sundermeyer et~al.(2021)Sundermeyer, Mousavian, Triebel, and Fox]{sundermeyer2021contact}
Martin Sundermeyer, Arsalan Mousavian, Rudolph Triebel, and Dieter Fox.
\newblock Contact-graspnet: Efficient 6-dof grasp generation in cluttered scenes.
\newblock In \emph{2021 IEEE International Conference on Robotics and Automation (ICRA)}, pages 13438--13444. IEEE, 2021.

\bibitem[Tang et~al.(2023)Tang, Lin, Akinola, Handa, Sukhatme, Ramos, Fox, and Narang]{tang2023industreal}
Bingjie Tang, Michael~A Lin, Iretiayo Akinola, Ankur Handa, Gaurav~S Sukhatme, Fabio Ramos, Dieter Fox, and Yashraj Narang.
\newblock Industreal: Transferring contact-rich assembly tasks from simulation to reality.
\newblock \emph{arXiv preprint arXiv:2305.17110}, 2023.

\bibitem[Tang et~al.(2024)Tang, Akinola, Xu, Wen, Handa, Van~Wyk, Fox, Sukhatme, Ramos, and Narang]{tang2024automate}
Bingjie Tang, Iretiayo Akinola, Jie Xu, Bowen Wen, Ankur Handa, Karl Van~Wyk, Dieter Fox, Gaurav~S Sukhatme, Fabio Ramos, and Yashraj Narang.
\newblock Automate: Specialist and generalist assembly policies over diverse geometries.
\newblock \emph{arXiv preprint arXiv:2407.08028}, 2024.

\bibitem[Team et~al.(2024)Team, Ghosh, Walke, Pertsch, Black, Mees, Dasari, Hejna, Kreiman, Xu, et~al.]{team2024octo}
Octo~Model Team, Dibya Ghosh, Homer Walke, Karl Pertsch, Kevin Black, Oier Mees, Sudeep Dasari, Joey Hejna, Tobias Kreiman, Charles Xu, et~al.
\newblock Octo: An open-source generalist robot policy.
\newblock \emph{arXiv preprint arXiv:2405.12213}, 2024.

\bibitem[Wang et~al.(2023{\natexlab{a}})Wang, Ling, Yuan, Shridhar, Bao, Qin, Wang, Xu, and Wang]{wang2023gensim}
Lirui Wang, Yiyang Ling, Zhecheng Yuan, Mohit Shridhar, Chen Bao, Yuzhe Qin, Bailin Wang, Huazhe Xu, and Xiaolong Wang.
\newblock Gensim: Generating robotic simulation tasks via large language models.
\newblock \emph{arXiv preprint arXiv:2310.01361}, 2023{\natexlab{a}}.

\bibitem[Wang et~al.(2022)Wang, Wu, Mo, Ke, Fan, Guibas, and Dong]{wang2022adaafford}
Yian Wang, Ruihai Wu, Kaichun Mo, Jiaqi Ke, Qingnan Fan, Leonidas~J Guibas, and Hao Dong.
\newblock Adaafford: Learning to adapt manipulation affordance for 3d articulated objects via few-shot interactions.
\newblock In \emph{European conference on computer vision}, pages 90--107. Springer, 2022.

\bibitem[Wang et~al.(2023{\natexlab{b}})Wang, Sun, Erickson, and Held]{wang2023one}
Yufei Wang, Zhanyi Sun, Zackory Erickson, and David Held.
\newblock One policy to dress them all: Learning to dress people with diverse poses and garments.
\newblock \emph{arXiv preprint arXiv:2306.12372}, 2023{\natexlab{b}}.

\bibitem[Wang et~al.(2023{\natexlab{c}})Wang, Xian, Chen, Wang, Wang, Fragkiadaki, Erickson, Held, and Gan]{wang2023robogen}
Yufei Wang, Zhou Xian, Feng Chen, Tsun-Hsuan Wang, Yian Wang, Katerina Fragkiadaki, Zackory Erickson, David Held, and Chuang Gan.
\newblock Robogen: Towards unleashing infinite data for automated robot learning via generative simulation.
\newblock \emph{arXiv preprint arXiv:2311.01455}, 2023{\natexlab{c}}.

\bibitem[Wu et~al.(2021)Wu, Zhao, Mo, Guo, Wang, Wu, Fan, Chen, Guibas, and Dong]{wu2021vat}
Ruihai Wu, Yan Zhao, Kaichun Mo, Zizheng Guo, Yian Wang, Tianhao Wu, Qingnan Fan, Xuelin Chen, Leonidas Guibas, and Hao Dong.
\newblock Vat-mart: Learning visual action trajectory proposals for manipulating 3d articulated objects.
\newblock \emph{arXiv preprint arXiv:2106.14440}, 2021.

\bibitem[Wu et~al.(2024)Wu, Wang, Guan, Jia, Liang, Song, Qu, Wang, Wang, Cao, et~al.]{wu2024fast}
Ziniu Wu, Tianyu Wang, Chuyue Guan, Zhongjie Jia, Shuai Liang, Haoming Song, Delin Qu, Dong Wang, Zhigang Wang, Nieqing Cao, et~al.
\newblock Fast-umi: A scalable and hardware-independent universal manipulation interface.
\newblock \emph{arXiv preprint arXiv:2409.19499}, 2024.

\bibitem[Xiang et~al.(2020)Xiang, Qin, Mo, Xia, Zhu, Liu, Liu, Jiang, Yuan, Wang, et~al.]{xiang2020sapien}
Fanbo Xiang, Yuzhe Qin, Kaichun Mo, Yikuan Xia, Hao Zhu, Fangchen Liu, Minghua Liu, Hanxiao Jiang, Yifu Yuan, He~Wang, et~al.
\newblock Sapien: A simulated part-based interactive environment.
\newblock In \emph{Proceedings of the IEEE/CVF conference on computer vision and pattern recognition}, pages 11097--11107, 2020.

\bibitem[Xiong et~al.(2024{\natexlab{a}})Xiong, Mendonca, Shaw, and Pathak]{xiong2024adaptive}
Haoyu Xiong, Russell Mendonca, Kenneth Shaw, and Deepak Pathak.
\newblock Adaptive mobile manipulation for articulated objects in the open world.
\newblock \emph{arXiv preprint arXiv:2401.14403}, 2024{\natexlab{a}}.

\bibitem[Xiong et~al.(2024{\natexlab{b}})Xiong, Varadarajan, Wu, Xiang, Xiao, Zhu, Dai, Wang, Sun, Iandola, et~al.]{xiong2024efficientsam}
Yunyang Xiong, Bala Varadarajan, Lemeng Wu, Xiaoyu Xiang, Fanyi Xiao, Chenchen Zhu, Xiaoliang Dai, Dilin Wang, Fei Sun, Forrest Iandola, et~al.
\newblock Efficientsam: Leveraged masked image pretraining for efficient segment anything.
\newblock In \emph{Proceedings of the IEEE/CVF Conference on Computer Vision and Pattern Recognition}, pages 16111--16121, 2024{\natexlab{b}}.

\bibitem[Xu et~al.(2022)Xu, He, and Song]{xu2022universal}
Zhenjia Xu, Zhanpeng He, and Shuran Song.
\newblock Universal manipulation policy network for articulated objects.
\newblock \emph{IEEE robotics and automation letters}, 7\penalty0 (2):\penalty0 2447--2454, 2022.

\bibitem[Xu et~al.(2023)Xu, Xian, Lin, Chi, Huang, Gan, and Song]{xu2023roboninja}
Zhenjia Xu, Zhou Xian, Xingyu Lin, Cheng Chi, Zhiao Huang, Chuang Gan, and Shuran Song.
\newblock Roboninja: Learning an adaptive cutting policy for multi-material objects.
\newblock \emph{arXiv preprint arXiv:2302.11553}, 2023.

\bibitem[Ze et~al.(2024)Ze, Zhang, Zhang, Hu, Wang, and Xu]{ze20243d}
Yanjie Ze, Gu~Zhang, Kangning Zhang, Chenyuan Hu, Muhan Wang, and Huazhe Xu.
\newblock 3d diffusion policy: Generalizable visuomotor policy learning via simple 3d representations.
\newblock In \emph{Robotics: Science and Systems}, 2024.

\bibitem[Zeng et~al.(2021)Zeng, Lee, Liang, and Kroemer]{zeng2021visual}
Vicky Zeng, Tabitha~Edith Lee, Jacky Liang, and Oliver Kroemer.
\newblock Visual identification of articulated object parts.
\newblock In \emph{2021 IEEE/RSJ International Conference on Intelligent Robots and Systems (IROS)}, pages 2443--2450. IEEE, 2021.

\bibitem[Zhang et~al.(2023)Zhang, Eisner, and Held]{zhang2023flowbot++}
Harry Zhang, Ben Eisner, and David Held.
\newblock Flowbot++: Learning generalized articulated objects manipulation via articulation projection.
\newblock \emph{arXiv preprint arXiv:2306.12893}, 2023.

\bibitem[Zhang et~al.(2024)Zhang, Liang, Chen, Ze, and Xu]{zhang2024catch}
Yuanhang Zhang, Tianhai Liang, Zhenyang Chen, Yanjie Ze, and Huazhe Xu.
\newblock Catch it! learning to catch in flight with mobile dexterous hands.
\newblock \emph{arXiv preprint arXiv:2409.10319}, 2024.

\bibitem[Zhao et~al.(2021)Zhao, Jiang, Jia, Torr, and Koltun]{zhao2021point}
Hengshuang Zhao, Li~Jiang, Jiaya Jia, Philip~HS Torr, and Vladlen Koltun.
\newblock Point transformer.
\newblock In \emph{Proceedings of the IEEE/CVF international conference on computer vision}, pages 16259--16268, 2021.

\bibitem[Zhou et~al.(2019)Zhou, Barnes, Lu, Yang, and Li]{zhou2019continuity}
Yi~Zhou, Connelly Barnes, Jingwan Lu, Jimei Yang, and Hao Li.
\newblock On the continuity of rotation representations in neural networks.
\newblock In \emph{Proceedings of the IEEE/CVF conference on computer vision and pattern recognition}, pages 5745--5753, 2019.

\bibitem[Zhu et~al.(2022)Zhu, Joshi, Stone, and Zhu]{zhu2022viola}
Yifeng Zhu, Abhishek Joshi, Peter Stone, and Yuke Zhu.
\newblock Viola: Imitation learning for vision-based manipulation with object proposal priors.
\newblock \emph{arXiv preprint arXiv:2210.11339}, 2022.
\newblock \doi{10.48550/arXiv.2210.11339}.

\bibitem[Zhuang et~al.(2023)Zhuang, Fu, Wang, Atkeson, Schwertfeger, Finn, and Zhao]{zhuang2023robot}
Ziwen Zhuang, Zipeng Fu, Jianren Wang, Christopher Atkeson, Soeren Schwertfeger, Chelsea Finn, and Hang Zhao.
\newblock Robot parkour learning.
\newblock \emph{arXiv preprint arXiv:2309.05665}, 2023.

\end{thebibliography}

\clearpage
\newpage
\appendix

\begin{figure}[t]
    \centering
    \includegraphics[width=\linewidth]{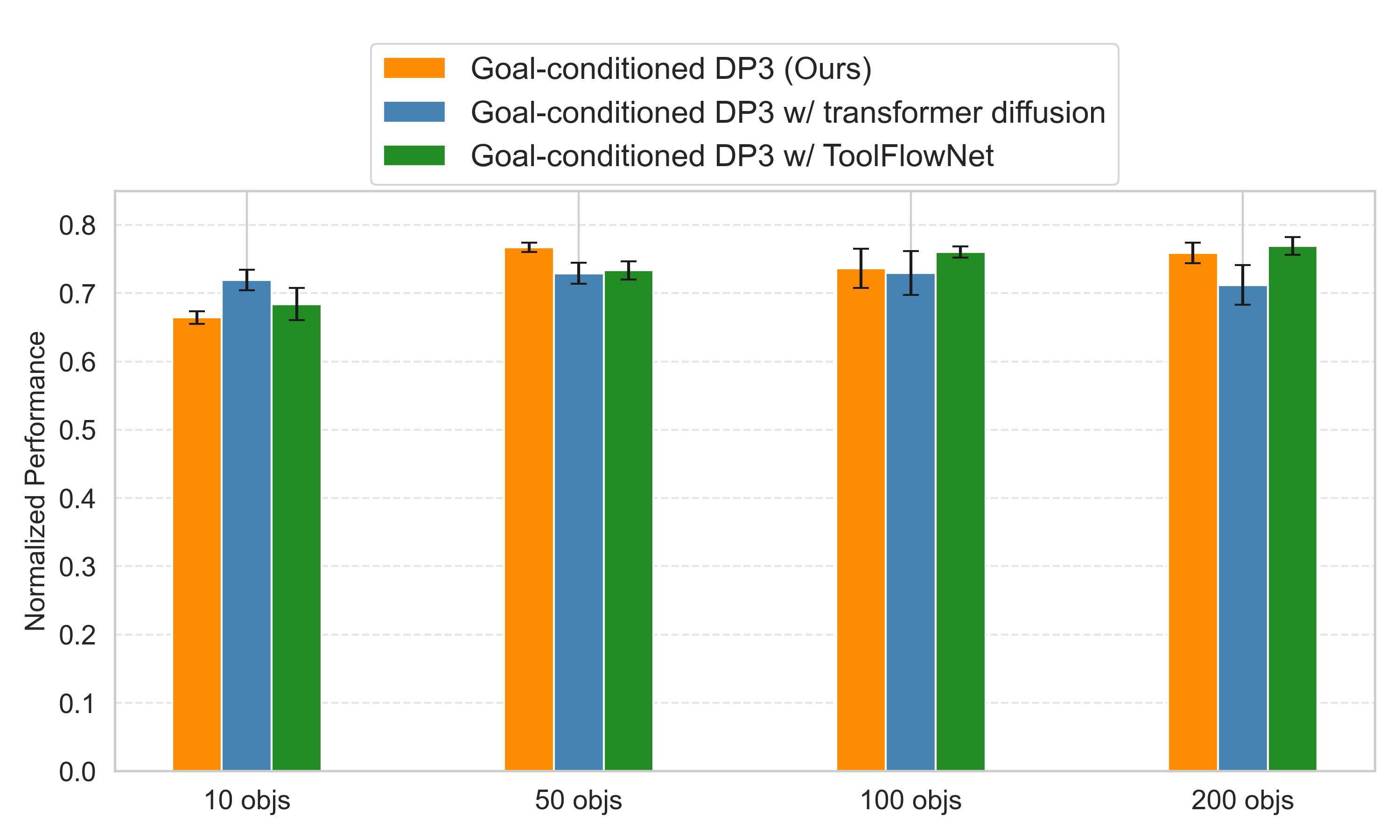}
    \vspace{-8mm}
    \caption{Comparison of different low-level policies.}
    \label{fig:low-level}
\end{figure}

\begin{figure*}[t]
    \centering
    \includegraphics[width=\linewidth]{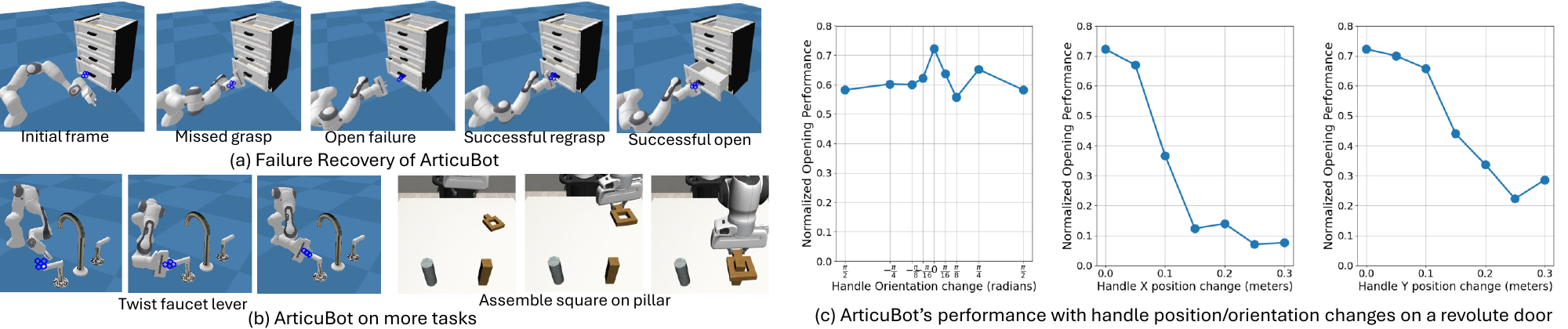}
    \caption{(a) Failure recovery ability of \method{}. (b) The proposed hierarchical policy learning framework can be applied to a broader range of manipulation tasks. (c) The performance of \method{} with augmented handles on a revolute door.}
    \label{fig:rebuttal}
\end{figure*}

\begin{figure*}[ht!]
\centering
    \includegraphics[width=.85\linewidth]{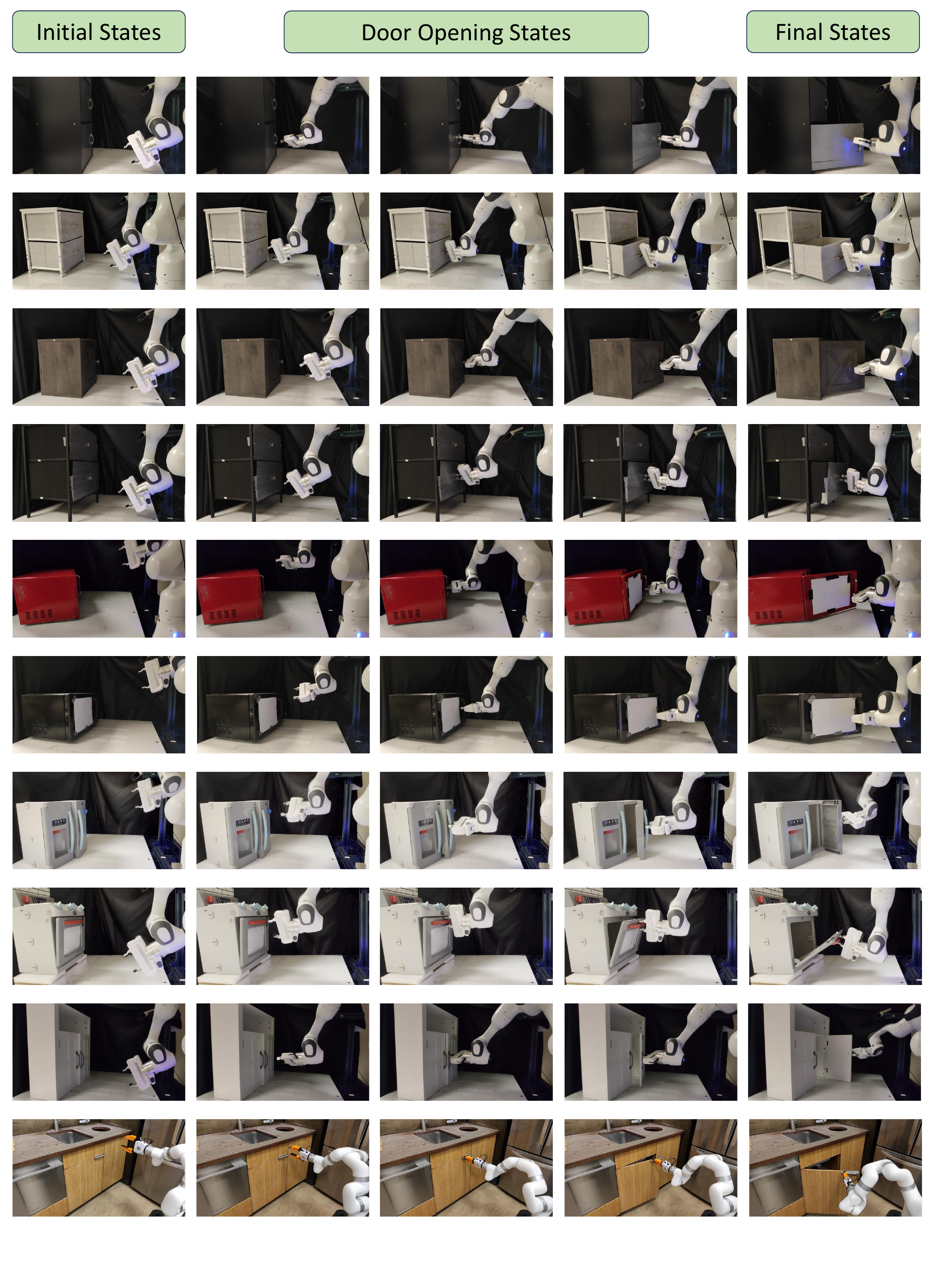}
    \vspace{-8mm}
    \caption{Rollouts of ArticuBot trying to open various unseen real-world articulated objects using a single learned policy across three robot setups in two different labs, as well as in real lounges and kitchens.
    }
    \vspace{-3mm}
    \label{fig:real_world_rollouts_1}
\end{figure*}

\begin{figure*}[ht!]
\centering
    \includegraphics[width=.85\linewidth]{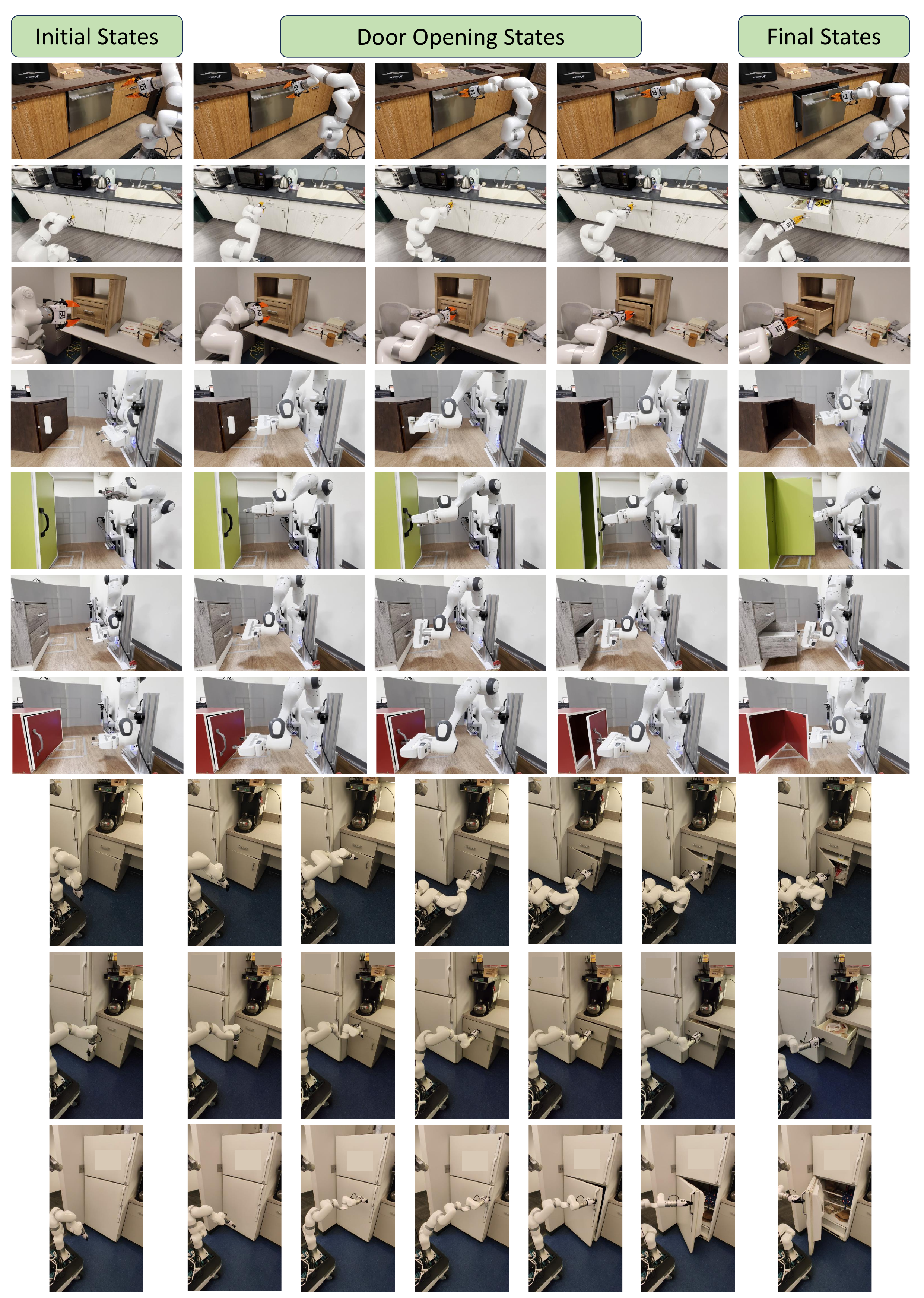}
    \caption{Rollouts of ArticuBot trying to open various unseen real-world articulated objects using a single learned policy across three robot setups in two different labs, as well as in real lounges and kitchens.
    }
    \vspace{-3mm}
    \label{fig:real_world_rollouts_2}
\end{figure*}

\subsection{Visualizations of \method{}'s performance in the real world}
\label{app:real_world_visual}

Fig.~\ref{fig:real_world_rollouts_1} and Fig.~\ref{fig:real_world_rollouts_2} provides visualizations of an opening trajectory by \method{} on all real-world test objects. 

\subsection{Visualizations of \method{}'s performance in simulation}

\begin{figure*}[ht!]
\centering
    \includegraphics[width=.8\linewidth]{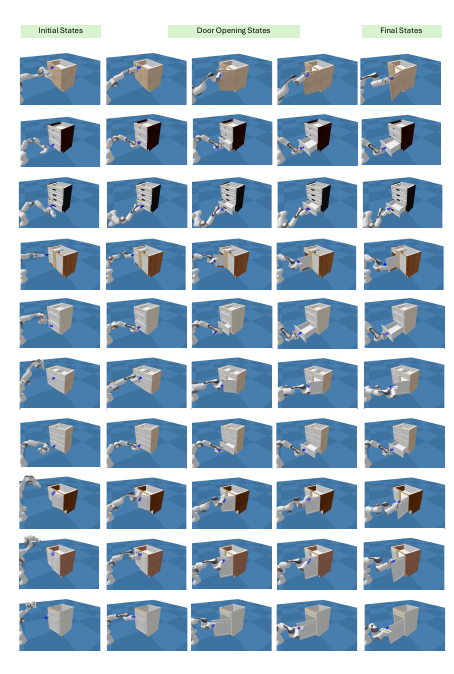}
    \vspace{-8mm}
    \caption{Rollouts of our learned high-level and low-level policy in simulation, on all 10 test objects. The 4 blue circles represent the predicted sub-goal end-effector points of the high-level policy.
    }
    \vspace{-3mm}
    \label{fig:append_our_method_success}
\end{figure*}

Fig.~\ref{fig:append_our_method_success} provides visualizations of a successful opening trajectory by \method{} on all 10 test objects in simulation.

\subsection{Comparison of different low-level policies}
\label{app:low-level-policy}
We test the performance of different low-level policy choices in this section. 
We compare our proposed \textbf{Goal-conditioned DP3} with UNet diffusion head to the following methods:
\begin{itemize}
    \item  \textbf{Goal-conditioned DP3 w/ transformer diffusion}: this baseline  replaces the UNet diffusion head in DP3 to be a transformer-based architecture. 
    \item  \textbf{Goal-conditioned DP3 w/ ToolFlowNet~\cite{seita2023toolflownet}} action space: instead of predicting the delta end-effector transformations as a 3D delta translation and a 6D delta rotation, it is represented as a 12D vector, where each 3 elements of the vector represents the 3D delta movement of one of the end-effector points $\{ee_i\}_{i=1}^4$. At test time, SVD is performed to extract the delta transformation of the end-effector using the predicted movements of the end-effector points. 
\end{itemize}
We use a fixed high-level policy (a weighted displacement model) for all experiments in this section.
We compare different methods trained without camera randomizations.
The results are shown in Fig.~\ref{fig:low-level}.
We do not observe huge performance differences between these different methods. We also do not observe huge benefits when increasing the amount of data beyond 50 objects. 
Our hypothesis is that, the goal end-effector pose provides a strong conditioning for the low-level policy; with such information as input, the differences in the policy architectures and action spaces may not matter too much. 




\subsection{Details of Simulation Randomization}
\label{app:simulation-randomization-details}
We randomize the pose and size of the object, and the initial joint angle of the Franka Arm in simulation when generating the demonstrations. 
The object's position is randomly sampled from: $x \sim [0.6, 1.2]$, $y \sim [-0.1, 0.1]$ and $z = 0$ (unit is meter). Its orientation along the z axis is randomly sampled from $[-30^\circ, 30^\circ]$. The size of the object (which is measured as the diagonal length of its 3D bounding box) is uniformly sampled from $[1, 1.8]$ meters. The robot joint angle is randomly initialized in simulation such that the end-effector is within 0.8 meters from the object.

\subsection{Hierarchical Policy Learning on More tasks}
\label{app:more-task}

We demonstrate that the proposed hierarchical policy learning framework works on more tasks. 
We tested faucet manipulation, where the robot grasps and twists the lever, achieving a normalized performance of 0.62 (trained on 50 demos). 
Additionally, on the MimicGen benchmark task of grasping a square nut and inserting it on a pillar, our method obtained a success rate of 0.60 (trained on 1000 demos). These experiments (see Figure~\ref{fig:rebuttal} (b)) show our approach's broader applicability to diverse manipulation tasks.

\subsection{Robustness of \method{} to Handle Augmentations}
\label{app:handle-augmentation}
We study the robustness of \method{} to changes in the position and orientations of the handle. 
To investigate this, we conducted an experiment systematically varying the handle's position and orientation (e.g., translating by 5 cm or rotating by 45 degrees) and analyzed the policy's performance. The results are shown in Figure~\ref{fig:rebuttal} (c). We find that the handle position affects the policy's performance more than the handle orientation.

\subsection{Implementation details of different high-level policies}
\label{app:high-level-detail}
In the main paper, we compared our Weighted Displacement Model with several other high-level policy representations: 
   \textbf{DP3 - UNet Diffusion}, 
   \textbf{DP3 - Transformer Diffusion}, and 
    \textbf{3D Diffuser Actor (3DDA)}~\cite{ke20243d}.
    We provide more details about these baselines here. 

For DP3 - UNet diffusion, the only change compared to the original DP3 is that we change the perception encoder from the default simplified PointNet encoder to be an attention-based encoder. This attention-based encoder is very similar to the one we use for the goal conditioned DP3 low-level policy as described in Sec.~\ref{sec:method-policy}. The only difference is that now the input to the policy does not have the sub-goal end-effector points, but just the scene point cloud and the current end-effector points. Therefore, we only perform cross attention between the scene points and the current end-effector points to obtain a latent embedding of the scene, which is then used as the condition in a U-Net diffusion head to generate the actions. 


The DP3 - Transformer Diffusion baselines is also modified on top of DP3. It has the same attention-based encoder as described above, which produces a latent embedding of the scene. 
Besides the change in the perception encoder, we also modify the UNet diffusion head in DP3 to be a transformer-based architecture, such that the diffusion head conditions on not only the latent embedding, but also the 3D point cloud features. 


Specifically, DP3 - transformer diffusion uses a transformer-based diffusion head to denoise the noisy version of the sub-goal end effector points $\{ee^{goal}_i\}_{i=1}^4$. 
The input to the transformer diffusion head includes observed current end-effector points $\{ee^{obs}_i\}_{i=1}^4$, the observed scene point cloud $P = \{p_j\}_{j=1}^M$, and the global feature, i.e., the latent embedding generated by the perception encoder $\mathcal{C}_{\text{global}}$. 
We also append the diffusion time step embedding to this global feature $\mathcal{C}_{\text{global}}$.
For both the observed end-effector points and the scene point cloud, a few history frames are appended with the current observation. However, for the sake of clarity, we ignore that in our notation.

The transformer-based diffusion head uses FiLM (Feature-wise Linear Modulation) conditioning~\cite{perez2018film} to process $\{ee^{obs}_i\}_{i=1}^4$, conditioned on the global feature $\mathcal{C}_{\text{global}}$, to generate the features for the observed end-effector points $\{ee^{obs\_feat}_i\}_{i=1}^4$. 
Similarly, FiLM-based global conditioning is applied to the scene points $\{p_j\}_{j=1}^M$, and the noisy sub-goal end-effector points $\{ee^{goal}_i\}_{i=1}^4$, conditioned on $\mathcal{C}_{\text{global}}$, to generate scene features $P^{feat}$ and the noisy sub-goal end-effector point features $\{ee^{goal\_feat}_i\}_{i=1}^4$, respectively.  

We then perform cross attention between the feature of the noisy sub-goal end-effector $\{ee^{goal\_feat}_i\}_{i=1}^4$ and the osberved end-effector $\{ee^{obs\_feat}_i\}_{i=1}^4$, followed by another cross attention between the output of the last cross attention and the scene feature \(P^{feat}\). 
We represent the output of the cross attention as \(\mathcal{F}_{ca}\). FiLM conditioning is again applied to \(\mathcal{F}_{ca}\), leveraging the global feature \(\mathcal{C}_{\text{global}}\), generating \(\mathcal{F}_{ca}^{feat}\). 
Finally, self attention is performed on \(\mathcal{F}_{ca}^{feat}\), and after which a MLP is applied to predict the noise.

For 3DDA, no global conditioning is used. Feed forward neural networks, i.e., MLPs, are used to generate scene features \(P^{feat}\) from the scene points $\{p_j\}_{j=1}^M$, and features for the noisy sub-goal end-effector $\{ee^{goal\_feat}_i\}_{i=1}^4$ from the noisy version of the sub-goal end effector points $\{ee^{goal}_i\}_{i=1}^4$, respectively. 
Cross attention is performed between the observed current gripper points, $\{ee^{obs}_i\}_{i=1}^4$ and the scene features \(P^{feat}\) to generate the observed gripper features, $\{ee^{obs\_feat}_i\}_{i=1}^4$. 
This is then encoded with diffusion timestep $t$ using feed forward layers to get the embedding \(t_{emb}\). 

Cross attention is then computed between the noisy goal end-effector point features $\{ee^{goal\_feat}_i\}_{i=1}^4$ and the scene points  features \(P^{feat}\), followed by Adaptive Layer Normalization between the output of the cross attention and \(t_{emb}\) to generate $\mathcal{F}_{ca}^{feat}$. 
Finally \(P^{feat}\) is concatenated with $\mathcal{F}_{ca}^{feat}$ and self attention is computed. To reduce computation complexity, farthest point sampling is applied on \(P^{feat}\) to reduce its dimensionality.

The main differences between DP3 - Transformer Diffusion and 3DDA  are as follows:
\begin{itemize}
    \item DP3 - Transformer uses many Film based global conditioning with the latent embedding, while 3DDA does not. 
    \item 3DDA uses self attention on the entire scene and the gripper points; DP3-transformer architecture performs only cross attention between the scene features and gripper features followed by self attention on the output. This significantly reduces the computation complexity of the network.
    \item 3DDA performs farthest point sampling on the scene features, while DP3 - Transformer does not.
\end{itemize}  






\subsection{Implementation details of prior articulated object manipulation methods}
\label{app:detail-prior-articulated-object}
\subsubsection{AO-Grasp~\cite{morlans2024grasp}}

AO-Grasp takes point clouds as input and generates grasp poses for articulate objects. It does not provide a policy to reach the required grasp pose or to open the object. 

To evaluate AO-Grasp in a similar fashion as our method, in simulation, we use Motion Planner to reach the target pose proposed by AO-Grasp, which gives us the "grasp accuracy". To calculate the "normalized opening performance", we use ground-truth articulation information of the object, available in simulation, to open the object. In the real world, we use our low-level policy to reach the target grasping pose produced by AO-Grasp. However, due to the lack of the ground-truth articulation information of the object, we cannot compare with AO-Grasp in terms of "normalized opening  performance" in the real world.

AO-Grasp is trained using point clouds in the camera frame. In the real world, we pass point clouds to the model in the camera frame, captured using Azure Kinects. In simulation, our point cloud observations are stored in the world frame, we first convert it to camera frame using the known camera extrinsics, and then use it as input to the AO-Grasp model to get the proposed grasps.

Finally, AO-Grasp returns a number of grasps sorted based on a score which is its belief on the correctness of the grasp. In cases where we have multiple doors (links) on the target object, we only keep the grasp proposals whose positions is near the target link to be opened. We then use a motion planner to reach the remaining grasp with the highest score. 

We observe that our method outperforms AO-Grasp in both simulation and real world. We attribute it to the following 2 reasons:

\begin{itemize}
    \item We find that most of the grasp proposals from AO-Grasp are at the edges and the corners of the doors. This may be because a large part of the AO-Grasp's training dataset are objects that are in partially opened states and grasping at the edges and the corners of the doors are valid grasps. However, in our test dataset, most of the doors are either closed or not opened enough to be opened using a grasp at the edge or the corner. 

    \item  Further, AO-Grasp generates many grasps at "fake" non-graspable edges resulted from the partial observation of the point cloud, which are not real graspable edges. 
\end{itemize}

\begin{figure*}[ht!]
\centering
    \includegraphics[width=\linewidth]{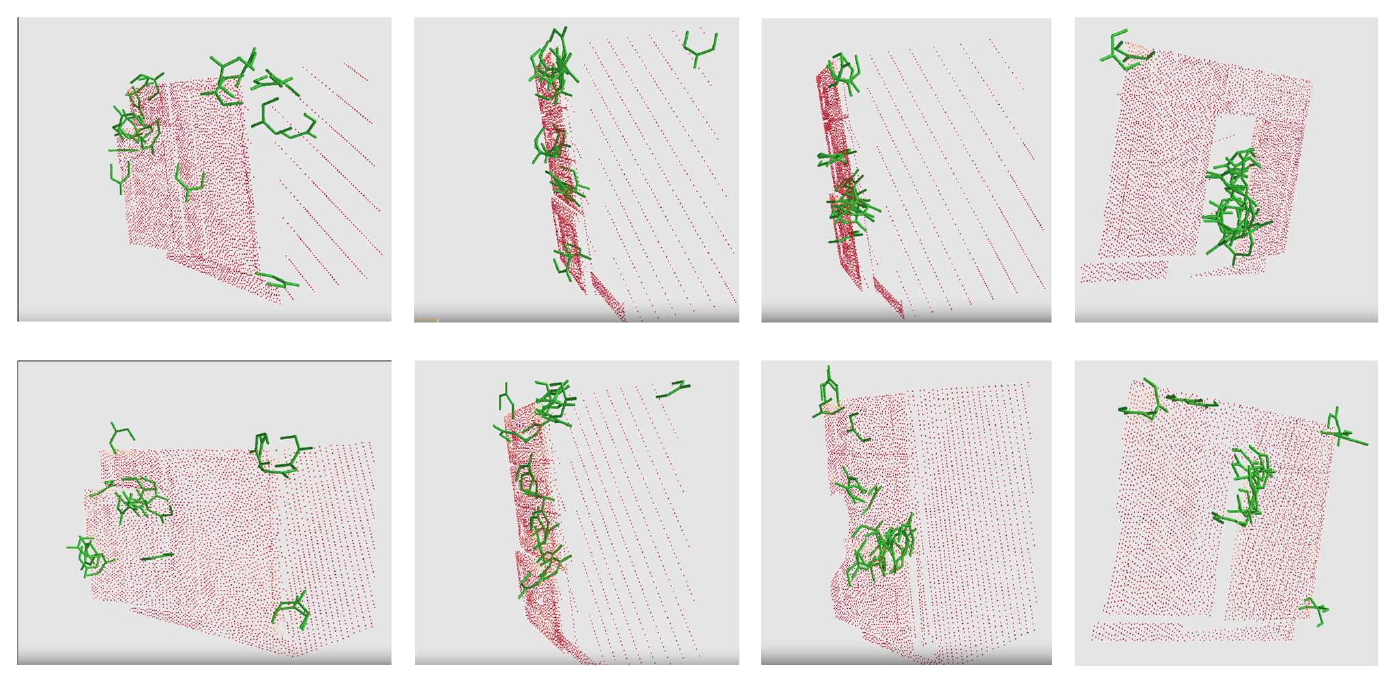}
    \vspace{-8mm}
    \caption{Visualization of AO-Grasp's failure cases in simulation. The green grippers represent the proposed grasps. For these examples, none of the grasps are executable by a motion planner. It can be observed that many of the grasps focus on fake edges of the object caused by the partial observation of the camera, which are not actually graspable. }
    \vspace{-3mm}
    \label{fig:append_ao_grasp_simulation_fail}
\end{figure*}

\begin{figure*}[ht!]
\centering
    \includegraphics[width=\linewidth]{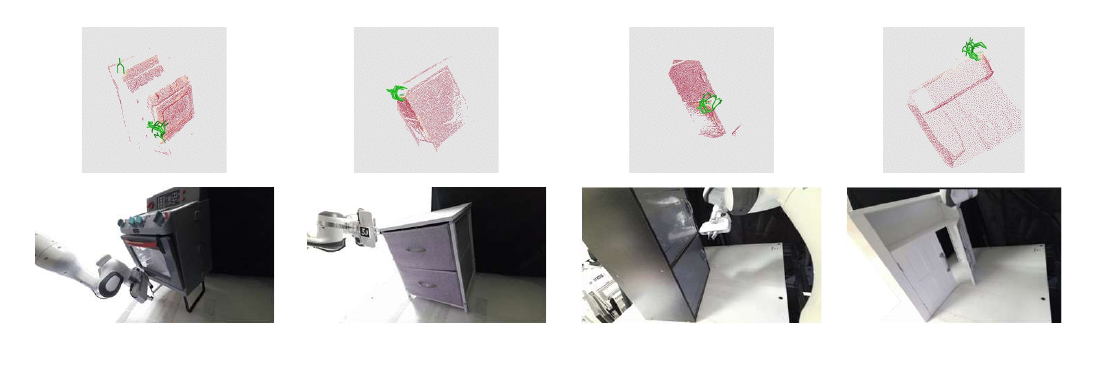}
    \vspace{-8mm}
    \caption{Visualization of the AO-Grasp failure cases in real world.  \textbf{Top}: the proposed grasps. \textbf{Bottom}: the execution result using our trained low-level policy on the highest score grasp. }
    \vspace{-3mm}
    \label{fig:append_ao_grasp_real_fail}
\end{figure*}

\begin{figure*}[ht!]
\centering
    \includegraphics[width=\linewidth]{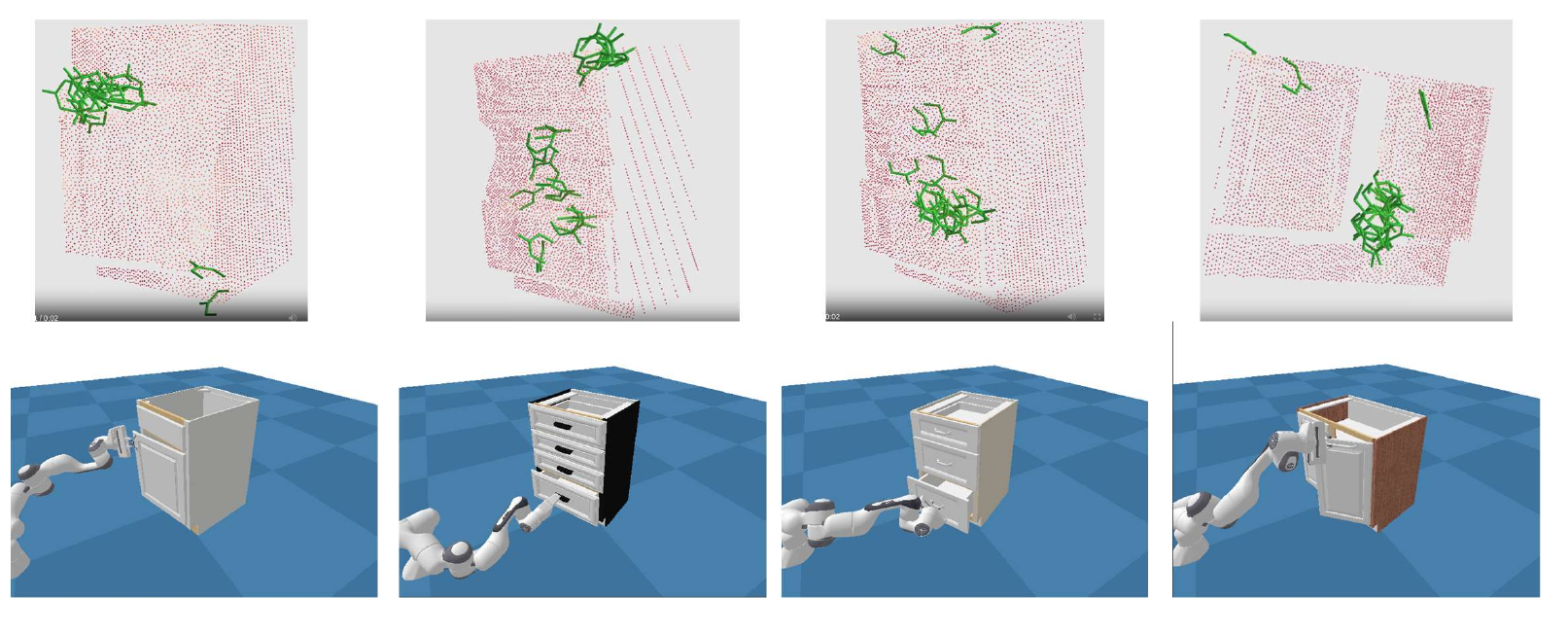}
    \vspace{-8mm}
    \caption{\textbf{Top}: Visualizations of AO-Grasp's proposed grasps (in green) on objects in simulation. \textbf{Bottom}: succesful grasps executed by the motion planner. }
    \vspace{-3mm}
    \label{fig:append_ao_grasp_simulation}
\end{figure*}

\begin{figure*}[ht!]
\centering
    \includegraphics[width=\linewidth]{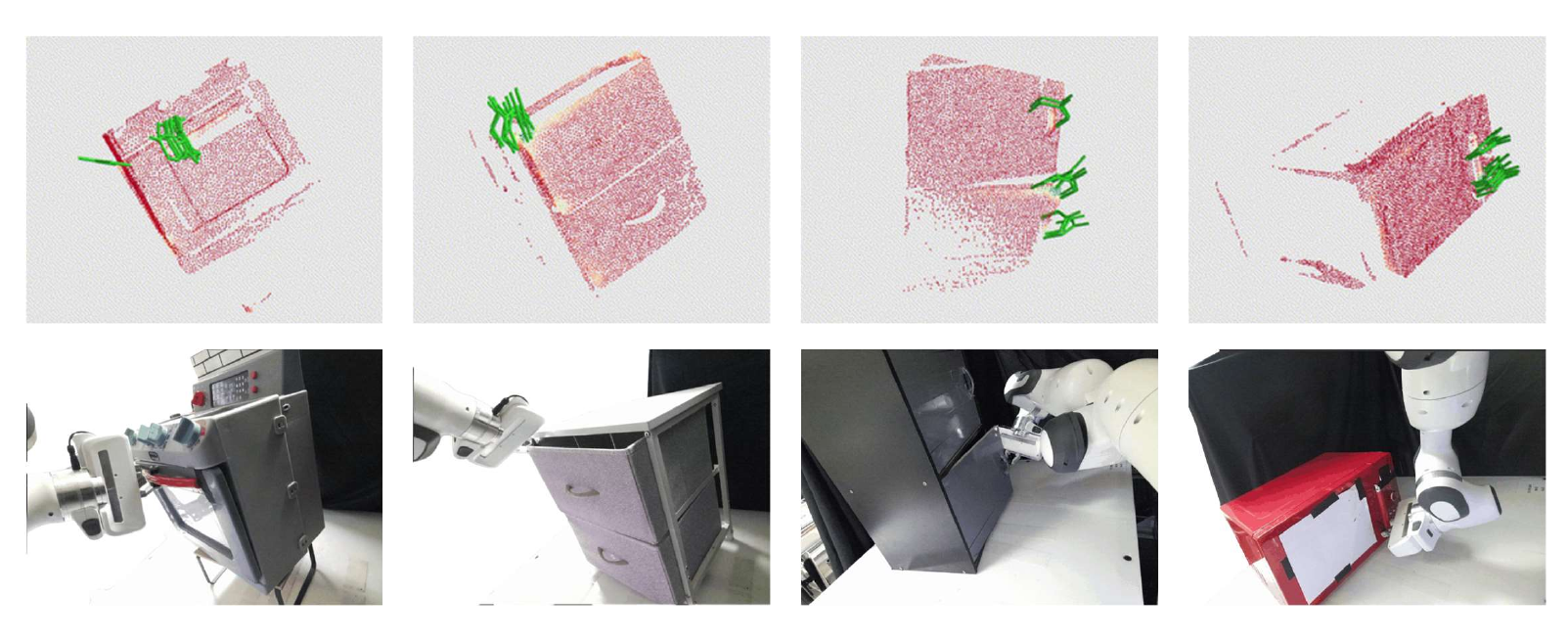}
    \vspace{-8mm}
    \caption{\textbf{Top}: Visualizations of AO-Grasp's proposed grasps (in green) on real-world object point clouds. \textbf{Bottom}: visuals of successful grasps  using \method's low-level policy to reach the generated grasp pose with the highest score. }
    \vspace{-3mm}
    \label{fig:append_ao_grasp_real}
\end{figure*}

In Fig.~\ref{fig:append_ao_grasp_simulation_fail} and Fig.~\ref{fig:append_ao_grasp_real_fail} we illustrate some of the failure grasps of AO-Grasps in simulation and real world, respectively. 
In Fig.~\ref{fig:append_ao_grasp_simulation} and Fig.~\ref{fig:append_ao_grasp_real}, we illustrate some of the successful grasps for AO-Grasp in simulation and real world respectively.


\subsubsection{FlowBot3D}
\textbf{Simulation:} FlowBot3D processes the point cloud of an articulated object and predicts the flow for each point on the object, guiding the robot’s end-effector to move along the direction of maximal flow to open the object. Additionally, it can incorporate an optional input in the form of a segmentation mask, which indicates whether a point belongs to the target link intended for opening, thereby enhancing performance. In our experiments, we use author-provided FlowBot3D checkpoints that were pre-trained on the PartNet-Mobility dataset, the same dataset employed for training and evaluation in our study.

The evaluation of FlowBot3D starts from the state where the robot gripper has firmly grasped the object handle. Subsequently, FlowBot3D is applied iteratively to open the door. 
At each step, the observation point cloud (along with the optional segmentation mask) is input into FlowBot3D to obtain the predicted flow for each point. 
From these predictions, the flow with the maximum magnitude within a 20 cm radius of the end-effector is selected as the direction for movement. The end-effector is then controlled to move by 2 cm in this direction, and the process is repeated 25 times to achieve the final result. 

\begin{figure*}[ht!]
\centering
    \includegraphics[width=\linewidth]{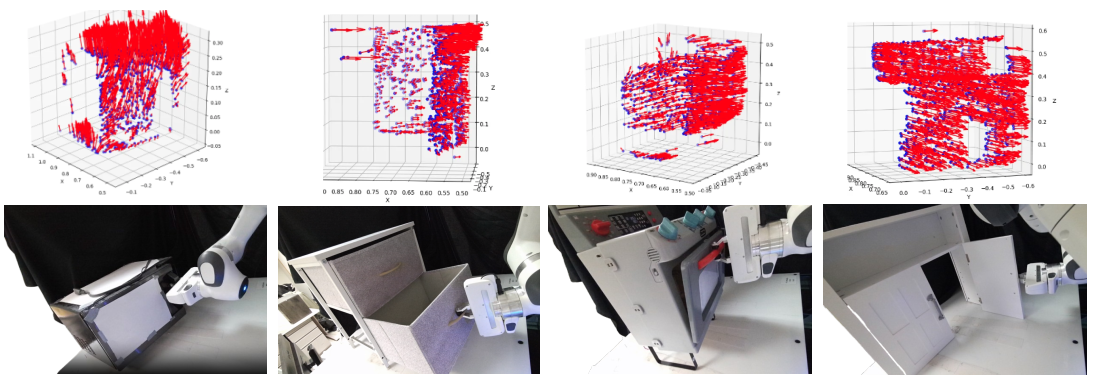}
    \caption{Visualizations of FlowBot3D's per-point flow. From left to right: 1. FlowBot3D predicts an upwards flow, causing the robot to attempt to lift the microwave. 2. The flow is parallel to the correct opening direction, resulting in a successful opening. 3. The flow direction is parallel to the ground instead of a downward circular arc, resulting in the oven getting dragged forward. 4. FlowBot3D correctly predicts a rightwards opening curve.}
    \vspace{-3mm}
    \label{fig:flowbot_prediction_real}
\end{figure*}

\textbf{Real World:} We initialize the scene by placing a single real-world test object on our table-top environment in Lab A. After manually moving the Franka end-effector in front of the object's handle, we close the gripper and ensure the handle is firmly grasped. 
At each timestep, observations from Azure Kinect cameras are fed through the perception pipeline outlined in Sec. \ref{app:perception-pipeline} to filter out unnecessary points and obtain a clean segmentation of the object point cloud. 
The processed point cloud is used to generate predicted flows from FlowBot3D. Following the set-up in simulation, the flow with the largest magnitude within a 20cm radius of the end-effector is used to determine the direction of the gripper action. The gripper is moved 2cm in the selected direction. This process is repeated up to 30 timesteps or until it is clear the articulated object will not open further, e.g. the robot experiences a torque limit overload, runs into joint limit, or the gripper slips from the handle.
See Fig.~\ref{fig:flowbot_prediction_real} for some successful and failure trials for FlowBot3D. 

\subsection{Implementation Details of Ablation Experiments in Sec.~\ref{sec:ablation}}
\label{app:detail-ablation}

\subsubsection{Weighted Displacement Model w/ Point Transformer}

In this ablation experiment, we substitute PointNet++ with the Point Transformer architecture of comparable size for the weighted displacement high-level policy. The Point Transformer leverages the transformer architecture specifically for point cloud processing, employing self-attention mechanisms to capture both local and global dependencies between points. 

 Here, similar to the PointNet++ model, the Point Transformer generates both the displacement to the sub-goal end-effector and a weight for each point in the observation point cloud. As demonstrated in Table~\ref{tab:ablation}, the performance of the Point Transformer is inferior to that of \method. 

\subsubsection{Unweighted Displacement Model}

In this ablation study, the model does not learn an individual weight for each point within the weighted displacement framework. Instead, the high-level policy predicts only the displacement to the sub-goal end-effector for each point. When generating the predicted sub-goal end-effector, the average of the predicted displacements across all points is computed. As demonstrated in Table~\ref{tab:ablation}, the omission of the weighting scheme results in a slight decline in performance. This underscores the importance of the weighting mechanism in emphasizing critical regions of the observation point cloud, such as the handle area, which are essential for achieving optimal results.

\subsubsection{Weighted Displacement Model with 6D orientation}

In our weighted displacement model, we predict the displacement from each point in the point cloud observation to the sub-goal end-effector and a weight for each point. 
Instead of predicting the displacement vectors to the goal end-effector points, this ablation learns to predict the following for each point in the scene:
1. The 3D displacement from current end-effector's position to the goal end-effector's positions. 
2. The absolute 6D orientation of the goal end-effector
3. The width between the fingers for the goal end-effector. 
The final prediction is still the weighted average prediction from each scene point.  

As shown in Table~\ref{tab:ablation}, the performance drops, suggesting that the proposed $4$ point representation of the sub-goal end-effector are both effective and essential. The suboptimal performance of the 6D orientation representation may stem from the ambiguity introduced by averaging the orientation and finger width outputs.

\subsubsection{Replacing low-level policy with a motion planner}

To demonstrate the effectiveness and necessity of our low-level policy, we conduct an ablation in which we replace the low-level policy with a motion planner and inverse kinematics (IK) controller. 
In the first stage of grasping the handle, we obtain the grasp pose by running our high-level policy on the initial observation of the system. We then use motion planning to move the end-effector to the predicted sub-goal. Following this, we close the gripper fingers to grasp the handle. In the second stage, which involves opening the door, we again run our high-level policy to predict the next sub-goal. Given the predicted sub-goal end-effector pose, we use IK to compute the joint angles required to reach the predicted sub-goal, and run the a PD controller to control the robot to that position.

The numerical results are presented in Table~\ref{tab:ablation}, which indicates that this approach performs poorly. A deeper analysis reveals several factors contributing to this suboptimal performance. The first issue is that the output of our high-level policy may not be sufficiently accurate, as it is applied only once at the beginning of each stage. 
For instance, since the predicted first sub-goal may be too close to or collide with the handle, motion planning fails to identify a collision-free path for the end-effector to reach the target position and grasp the handle. The second issue arises when directly applying the joint angles computed by IK to control the robot. This approach does not reliably guide the end-effector along a feasible path, such as a segment of a circle or a straight line, which is required for tasks like opening the door. Additionally, manually controlling the gripper to close and grasp the handle leads to issues where the gripper often slips off the handle due to an improper grasp and suboptimal trajectory.

These poor results highlight the crucial role of our low-level policy in effectively controlling the robot and ensuring the success of tasks such as grasping and manipulation.

\subsubsection{Additional ablation on gripper representations}

In our paper, we propose to represent the robot end-effector as a collection of 4 points, instead of a 3D position and a 6D orientation. In the previous ablation ``Weighted Displacement Model with 6D orientation'', we performed an experiment in which the weighted displacement model is learned to predict the 6D orientation of the goal end-effector instead of the displacements to the goal end-effector points, and it leads to much worse performance. One possible reason could be that it is hard to correctly average multiple predicted 6D orientations in the weighted displacement model. 

Therefore, here we compare to an additional ablation where we use the DP3 - UNet diffusion policy architecture (for the high-level policy, and the goal-conditioned low-level policy), and the end-effector is represented using the 3D position, 6D orientation, and 1D finger openness (we term this the 10D representation), instead of diffusing the 4 goal end-effector points. 
This policy architecture does not involve the average of multiple 6D rotations; thus it is a more controlled experiment to solely compare the 4 point end-effector representation and the 10D representation of the end-effector. 
In detail, we replace the representation of the current end-effector, and the goal end-effector, all from the 4 point representation to the 10D representation. We train a new high-level DP3 - UNet diffusion policy and a low-level goal-conditioned DP3 policy using this representation. 

We find that when using the 10D representation to train only a low-level policy, i.e., combining it with the high-level weighted displacement policy with the 4 point end-effector representation, the performance degrades from $0.7$ to $0.664$.
When using the 10D representation for both the high-level policy and the low-level policy, the performance further degrades to $0.465$.
This validates the effectiveness of the 4 point end-effector representation.



\subsection{Point cloud augmentations in simulation}
\label{app:point-cloud-augmentation}
We follow \cite{dalal2024local} to add two types of augmentations to the depth map rendered in simulation to mimic the noise in real-world depth cameras. 
The first is edge artifacts that models the noise along object edges. Specifically, we introduce correlated depth noise using bilinear interpolation on a shifted depth map. Given a depth map of size \( H \times W \), we first construct a new grid \(\{0, \dots, H-1\} \times \{0, \dots, W-1\}\). Each depth value in the grid is randomly shifted by a value sampled from \( N(0, 0.5) \) with a probability of 0.8. Finally, we apply bilinear interpolation between the original depth values and the shifted depth maps to generate the new depth map.

The second is random holes in the depth map to model random depth pixel value loss in real-world depth cameras. We generate a random pixel-level mask from \( U(0,1) \), smooth it with Gaussian blur, and normalize it to \([0,1]\). Pixels with mask values exceeding a randomly sampled threshold from \( U(0.6,0.9) \) are set to zero. This randomization is applied to a depth map with a probability of 0.5.

In the real world, we also use statistical and radius outlier filters to remove outliers in the point cloud (details in Sec.~\ref{app:perception-pipeline}). We also apply such filters to the simulated point cloud after the above two augmentations, such that the point cloud in simulation is made to be similar as the ones we obtain in the real world as much as possible. The standard deviation ratio used in statistical outlier remover is randomly sampled from $[0.4, 0.6]$, and the number of neighbors used in the radius outlier remover is randomly sampled from $[20, 95]$.

\subsection{Real-world perception pipeline details}
\label{app:perception-pipeline}
Unlike in simulation, depth maps obtained from real-world depth sensors are usually noisy, e.g., there are random holes in the depth maps, and depth value interpolations along object edges. Therefore, the depth maps need to be processed to extract the point cloud of the target articulated object. 
We use two slightly different perception pipelines for extracting object point clouds during the grasping and opening phase of each trial.

\textbf{Grasping Phase:} 
We first obtain RGB images and point clouds from depth cameras (Azure Kinects in Lab A and Mobile X-Arm, and RealSense D-435s in lab B). 
As shown in Fig. \ref{fig:real_world_setup}, a camera is placed on each side of the robot arm in all setups (in total two cameras are used). 
We assume that in the first frame, the robot arm is not occluding the target object in the camera. Based on this assumption, GroundingDino~\cite{liu2025grounding} and EfficientSAM~\cite{xiong2024efficientsam} are used to obtain a clean segmentation mask of the object of interest. We use simple text prompts to detect the object, such as \textit{microwave} and \textit{drawer}.

In addition, we obtain a cleaner object point cloud by explicitly segmenting out the robot arm from the point cloud. 
To do so, we first generate a canonical point cloud of the manipulator arm from the robot urdf and mesh files, and transform the robot point cloud using the current joint angles of the robot. 
We use camera-to-robot transformations and camera intrinsics to project the transformed robot arm point cloud to the 2D image. 
The resulting robot mask is relatively sparse, so we use the OpenCV dilation function to generate a denser mask of the robot arm. 
All pixels in the dilated robot mask are removed when the object-only point cloud is obtained.

Once the object point cloud from each camera is obtained, we use off-the-shelf point cloud processing libraries (Open3D) to clean the point cloud in the following steps:

\begin{enumerate}
\item Voxel Downsample: We downsample the point cloud by voxelizing it into 2mm voxels. This allows us to run the latter processing steps more efficiently. 
\item Radius Outlier Removal: Each point is required to have at least 20 neighboring points within a radius of 2 cm. If this criterion is not met, the point is removed.
\item Statistical Outlier Removal: For each point in the point cloud, we calculate the average distance from itself to 20 of its closest neighbors. We also compute the standard deviation of the average distances for all points in the whole point cloud. Any point whose average distance to its neighbors that is larger than 0.5 of the standard deviation is removed.  

\end{enumerate}

The final point cloud is obtained by merging the point clouds from the two cameras using known calibration transformations. We then perform farthest point sampling to obtain an object point cloud of 4500 points (which is the number of points in our simulation generated point clouds that the policy is trained on).


\textbf{Opening Phase:} Due to occlusions from the robot arm, the articulated object is more difficult to be cleanly detected and segmented out during the opening phase of the task. Instead, we use the point cloud obtained from the grasping phase to calculate a 3D bounding box of the articulated object. 
Then, this 3D bounding box is expanded to include the end-effector position, since the opened link of the object will always be in front of the end-effector. 
We still remove the robot points and perform the same downsampling and outlier removal steps from the grasping phase to clean the resulting point cloud. Again, farthest point sampling is used to extract a representative point cloud (4500 points) of the object. 

\subsection{Failure cases of \method{} in the real world}
\label{app:failure-case}
\begin{figure*}
    \includegraphics[width=\textwidth]{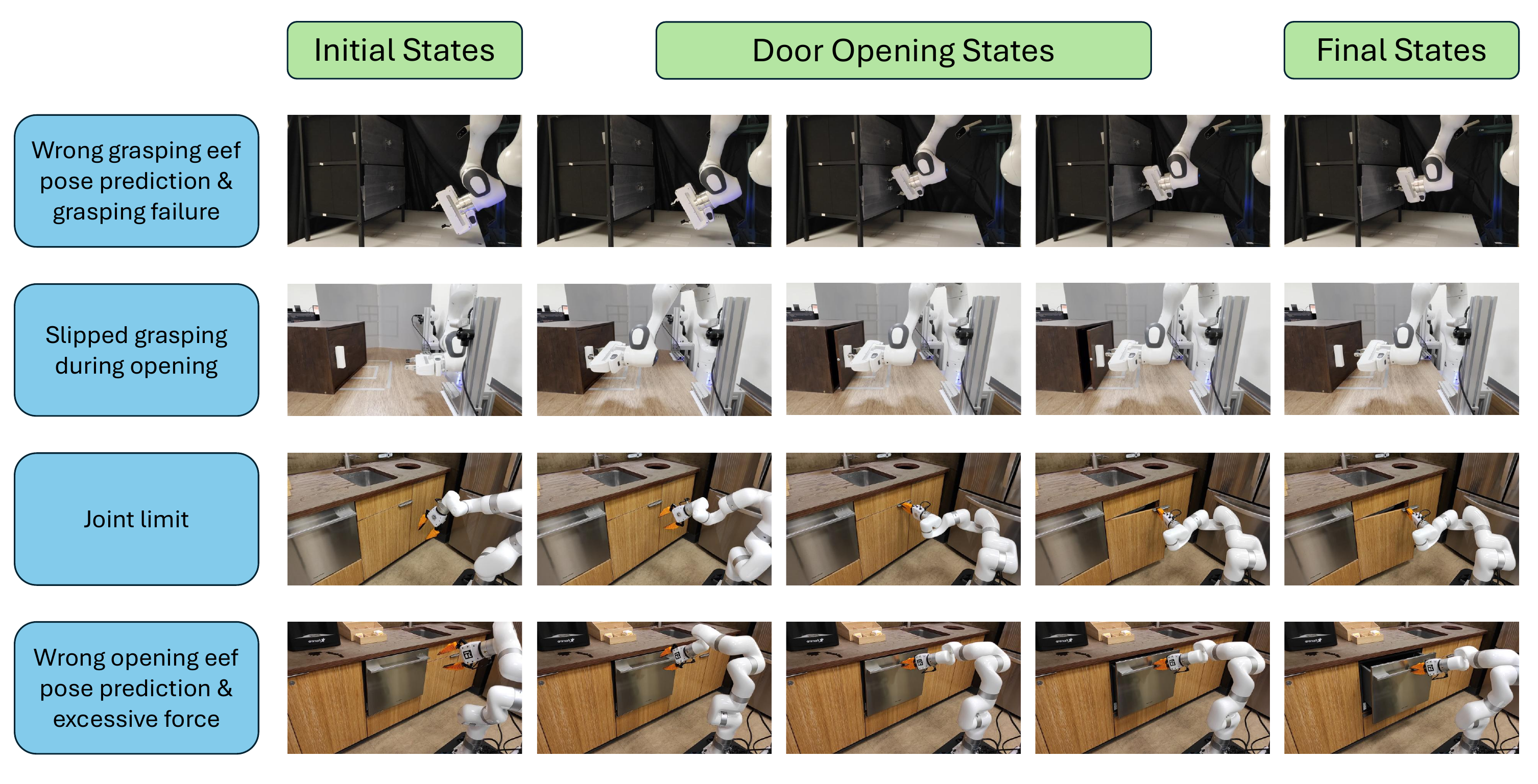}
    \caption{Visualizations of some of the failure cases of \method{} on some of the real-world tested objects. 
    \textbf{First row:} the high-level policy's predicted end-effector pose for grasping the handle is not accurate enough, leading to a grasp failure.
    \textbf{Second row:} the robot end-effector did not form a stable enough grasp of the handle, and it slipped off the handle partway during the opening.
    \textbf{Third row:} the robot runs into joint limit while opening the cabinet.
    \textbf{Fourth row:} the high-level policy predicts the opening end-effector pose as opening downwards, as many dishwashers have a revolute joint and open downwards. However, this dishwasher has a prismatic joint and pulls out horizontally. The low-level policy then tries to move the robot end-effector moving downwards, which is the wrong opening direction, resulting in excessive force on the robot arm and the arm is forced to stop for motor prediction. 
    }
    \label{fig:append_failure}
\end{figure*}

Fig.~\ref{fig:append_failure} shows some of the failure cases of \method{} in the real world. Please refer to the figure captions for detailed descriptions of the failure reasons. 
In Fig.~\ref{fig:rebuttal} (a) we show that \method{} has basic trajectory recovery capabilities during grasping or opening failures. This recovery behavior arises from two factors: (1) ArticuBot is a closed-loop system, continuously running both the high-level and low-level policies at every time step; thus, our method can correct its trajectory after encountering failures; and (2) we generated a large demonstration dataset that partially covers common failure states, enabling the method to recover from failures. 
However, such recovery abilities remain limited when failures occur in states not covered by our demonstrations. We leave exploring methods like DAgger to improve robustness as future work.

\end{document}